\documentclass[acmtog]{acmart}

\AtBeginDocument{%
  }

\setcopyright{acmlicensed}
\copyrightyear{2025}
\acmYear{2025}
\acmDOI{XXXXXXX.XXXXXXX}

\makeatletter
\DeclareRobustCommand\onedot{\futurelet\@let@token\@onedot}
\def\@onedot{\ifx\@let@token.\else.\null\fi\xspace}

\makeatother

\newcommand{\ignore}[1]{}   % ignore this
\newcommand{\topic}[1]
{
\vspace{1mm}\noindent\textbf{#1}
}
\graphicspath{{figures}, {examples}}
\usepackage{xcolor}
\colorlet{dark-blue}{blue!50!black}
\colorlet{dark-cyan}{cyan!75!black}
\colorlet{dark-purple}{purple!50!black}
\colorlet{dark-red}{red!75!black}
\colorlet{dark-green}{green!80!black}
\colorlet{dark-orange}{orange!50!black}
\colorlet{dark-gray}{black!75}
\colorlet{light-gray}{black!30}
\definecolor{nice-red}{HTML}{E41A1C}
\definecolor{nice-orange}{HTML}{FF7F00}
\definecolor{nice-yellow}{HTML}{FFC020}
\definecolor{nice-green}{HTML}{39b54a}
\definecolor{nice-blue}{HTML}{0071bc}
\definecolor{nice-purple}{HTML}{984EA3}

\colorlet{verylight-gray}{black!10}
\definecolor{LightCyan}{rgb}{0.66,0.85,0.76}

\usepackage{float}
\usepackage{lscape}                                         % Useful for wide tables or figures.
\usepackage{wrapfig}
\usepackage{tikz}
\usepackage{pgfplots}
\pgfplotsset{compat=1.18}

\usepackage[lined,ruled,linesnumbered]{algorithm2e}
\usepackage{animate} %% convert frames to video

\usepackage{booktabs}                   % Publication quality tables
\usepackage{multirow}

\usepackage{makecell}

\usepackage{paralist}
\usepackage{enumitem}
\setlist[enumerate]{itemsep=0mm}

\usepackage{bm}                          % Make bold, italic math symbols
\usepackage{epsfig}                      % for figures

\usepackage{bm}
\usepackage{mathtools}

\usepackage{amssymb,amsmath}   % Short math guide for LaTeX ftp://ftp.ams.org/pub/tex/doc/amsmath/short-math-guide.pdf

\usepackage{units}
\usepackage{color}

\usepackage{comment}

\usepackage{hyperref}
\usepackage{xcolor}
\begin{document}

\title{Magic Fixup: Streamlining Photo Editing by Watching Dynamic Videos}

\author{Hadi Alzayer}
\email{hadi@umd.edu}
\orcid{0000-0002-3215-5678}
\affiliation{%
  \institution{University of Maryland \& Adobe}
  \city{College Park}
  \state{MD}
  \country{USA}
}

\author{Zhihao Xia}
\email{zhihao.zach.xia@gmail.com}
\orcid{0000-0003-1548-2113}
\affiliation{%
  \institution{Adobe}
  \city{San Jose}
  \state{CA}
  \country{USA}
}

\author{Xuanar Zhang}
\email{cezhangxer@gmail.com}
\orcid{0000-0002-7679-800X}
\affiliation{%
  \institution{Adobe}
  \city{San Jose}
  \state{CA}
  \country{USA}
}

\author{Eli Shechtman}
\email{elishe@adobe.com}
\orcid{0000-0002-6783-1795}
\affiliation{%
  \institution{Adobe}
  \city{Seattle}
  \state{CA}
  \country{USA}
}

\author{Jia-Bin Huang}
\email{jbhuang@umd.edu}
\orcid{0000-0002-0536-3658}
\affiliation{%
  \institution{Adobe}
  \city{San Francisco}
  \state{CA}
  \country{USA}
}

\author{Michael Gharbi}
\email{mgharbi@gmail.com}
\orcid{0000-0002-4190-6955}
\affiliation{%
  \institution{Adobe}
  \city{San Francisco}
  \state{CA}
  \country{USA}
}

\ccsdesc[500]{Computing methodologies~Image editing}


\keywords{Photorealistic editing, Spatial editing, Learning from videos}


\newcommand{\orange}[1]{{\color{orange}#1}} % R2
\definecolor{wildstrawberry}{rgb}{1.0, 0.26, 0.64}% R3
\newcommand{\myparagraph}[1]{\vspace{2mm} \noindent \textit{#1}}
\newcommand{\hadi}[1]{{\textcolor{red}{[\emph{{Hadi}: #1}]}}}
\newcommand{\mg}[1]{{\textcolor{blue}{[\emph{{michael}: #1}]}}}
\newcommand{\new}[1]{{{#1}}}
\newcommand{\first}[1]{{\textcolor{cyan}{[\emph{{R1}: #1}]}}}
\newcommand{\secondr}[1]{\orange{[\emph{{R2}: #1}]}}
\newcommand{\third}[1]{{\textcolor{violet}{[\emph{{R3}: #1}]}}}
\newcommand{\fourth}[1]{{\textcolor{wildstrawberry}{[\emph{{R4}: #1}]}}}
\definecolor{hotpink}{RGB}{255, 20, 147}

\begin{teaserfigure}
% \vspace{-9mm}
\includegraphics[width=\linewidth]{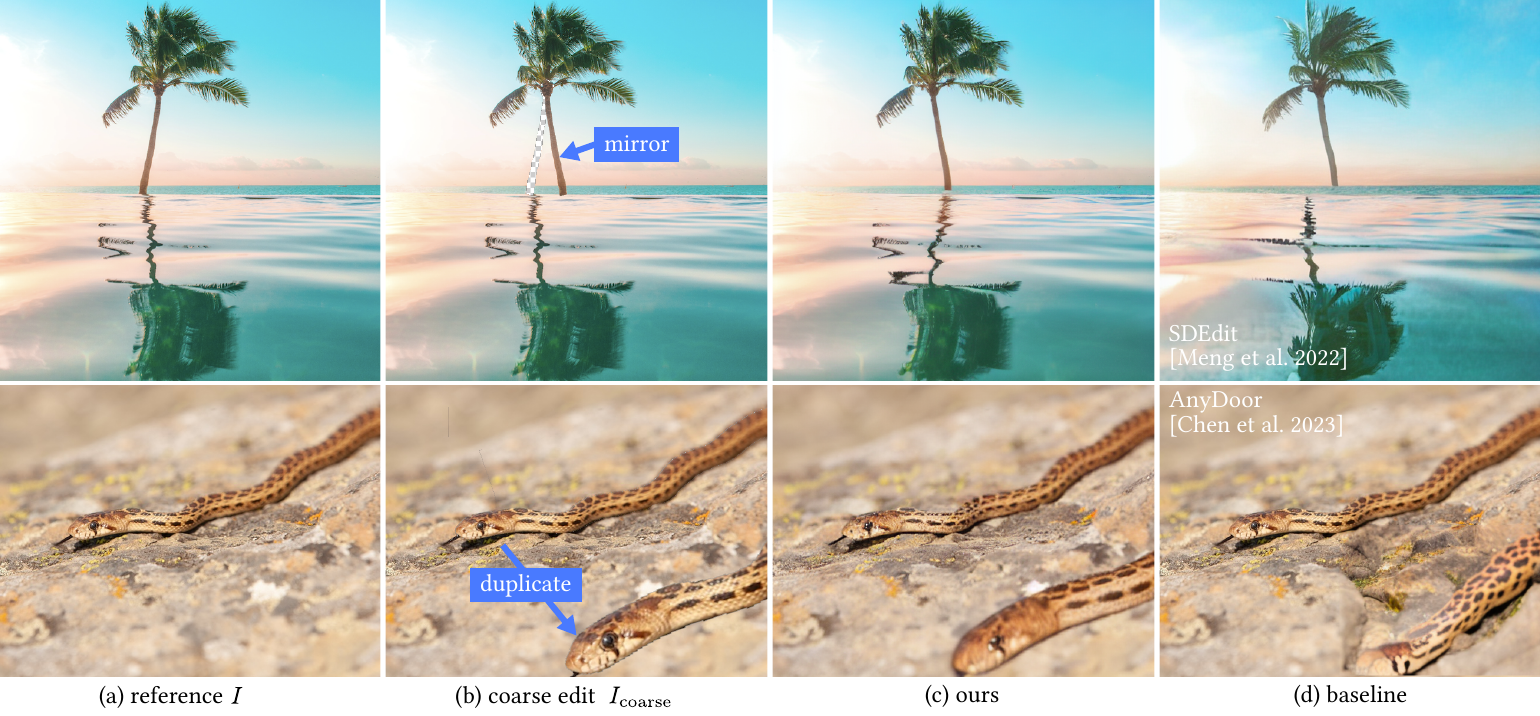}
% \vspace{-5mm}
\captionof{figure}{\label{fig:teaser}
\textbf{Applications of Magic Fixup.} We propose a diffusion model for image editing.
Starting from an input image (a), a user specifies their desired changes by rearranging automatically segmented scene objects using simple 2D transforms to produce a coarse edit (b).
Our model transforms this coarse edit into a realistic image (c), correctly accounting for secondary effects critical for realism, such as reflections on the water (top) or changes in depth-of-field (bottom), producing much more plausible edits than state-of-the-art methods (d). Photos sourced from \copyright Unsplash.
}
% \vspace{-9mm}
\end{teaserfigure}

% % \vspace{-9mm}
% \includegraphics[width=\linewidth]{figure_imgs/diverse_applications_optimized2.png}
% % \vspace{-5mm}
% \captionof{figure}{\label{fig:teaser}
% \new{
% \textbf{MagicFixup applications.} Our method can generalize to a large number of different applications. Here we show a selection of tasks we used MagicFixup to achieve, like image recomposition, perspective editing, reposing, 3D transformation, and colorization.
% }
% }
% % \vspace{-9mm}

\begin{abstract}
\begin{teaserfigure}
% \vspace{-9mm}
\includegraphics[width=\linewidth]{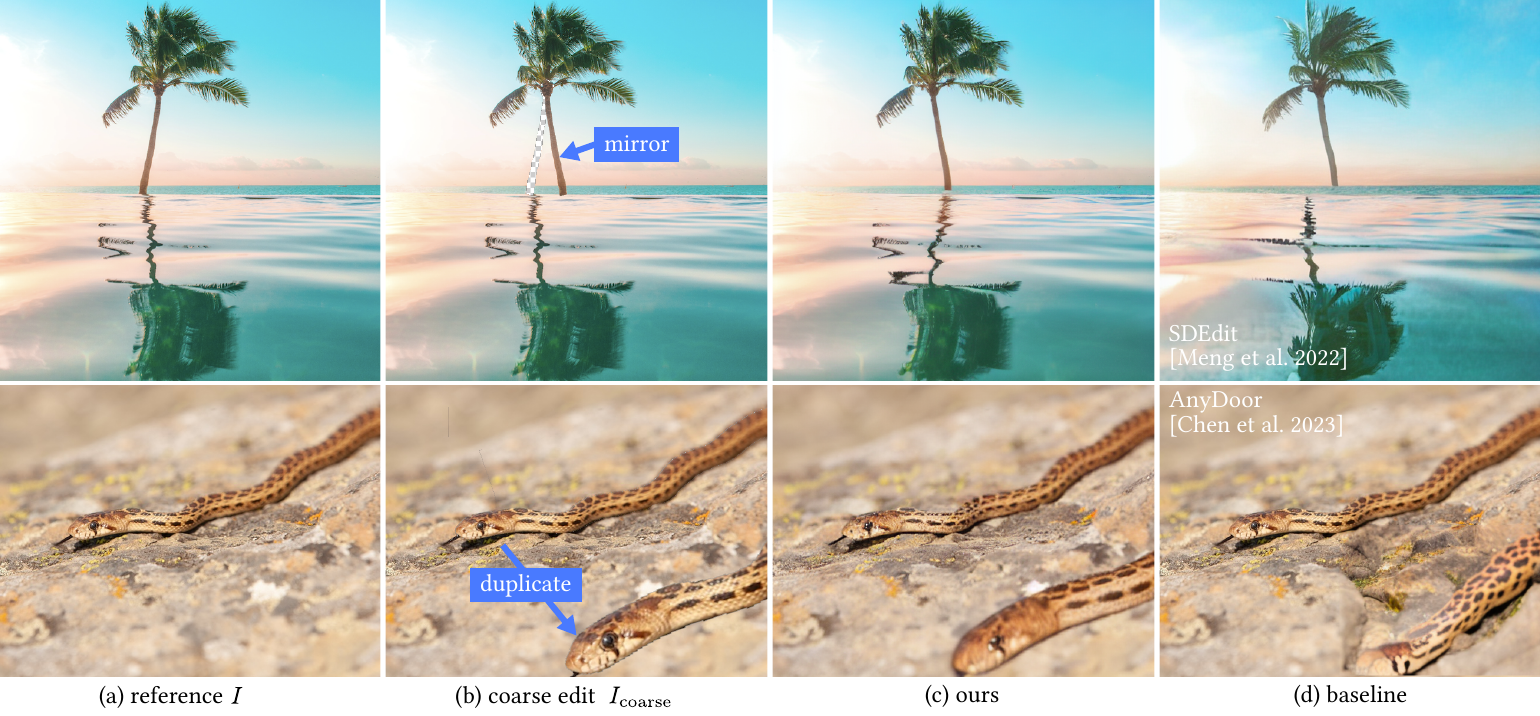}
% \vspace{-5mm}
\captionof{figure}{\label{fig:teaser}
\textbf{Applications of Magic Fixup.} We propose a diffusion model for image editing.
Starting from an input image (a), a user specifies their desired changes by rearranging automatically segmented scene objects using simple 2D transforms to produce a coarse edit (b).
Our model transforms this coarse edit into a realistic image (c), correctly accounting for secondary effects critical for realism, such as reflections on the water (top) or changes in depth-of-field (bottom), producing much more plausible edits than state-of-the-art methods (d). Photos sourced from \copyright Unsplash.
}
% \vspace{-9mm}
\end{teaserfigure}

% % \vspace{-9mm}
% \includegraphics[width=\linewidth]{figure_imgs/diverse_applications_optimized2.png}
% % \vspace{-5mm}
% \captionof{figure}{\label{fig:teaser}
% \new{
% \textbf{MagicFixup applications.} Our method can generalize to a large number of different applications. Here we show a selection of tasks we used MagicFixup to achieve, like image recomposition, perspective editing, reposing, 3D transformation, and colorization.
% }
% }
% % \vspace{-9mm}

% Editing a photograph by rearranging its components to produce a realistic output is a meticulous and time-consuming process prone to illusion-breaking mistakes. 
%
% However, making rough edits to convey the intended composition and concept quickly is easy for a human. 
%
We propose a generative model that, given a coarsely edited image, synthesizes a photorealistic output that follows the prescribed layout. 
Our method transfers fine details from the original image and preserve the identity of its parts.
Yet, it adapts it to the lighting and context defined by the new layout. 
Our key insight is that videos are a powerful source of supervision for this task: objects and camera motions provide many observations of how the world changes with viewpoint, lighting, and physical interactions. 
We construct an image dataset in which each sample is a pair of source and target frames extracted from the same video at randomly chosen time intervals. 
We warp the source frame toward the target using two motion models that mimic the expected test-time user edits. 
We supervise our model to translate the warped image into the ground truth, starting from a pretrained diffusion model.
% and fine-tune it to explicitly adhere to both the user input and the original image. 
%
Our model design explicitly enables fine detail transfer from the source frame to the generated image,
while closely following the user-specified layout.
% of our training procedure. 
%
We show that by using simple segmentations and coarse 2D manipulations, we can synthesize a photorealistic edit faithful to the user's input while addressing second-order effects like harmonizing the lighting and physical interactions between edited objects.
Project page and code can be found at {\color{hotpink}\url{https://magic-fixup.github.io}}
\end{abstract}

\maketitle

% TODO UNCOMMENT
\section{Introduction}
\label{sec:intro}
% What the problem is?

% Why is it important? 

% Why is it hard? 

% What other people have done to address it?

% Why do the existing methods not satisfactory?

% What we have done? Start with ``In this paper, "
% - Provide forward references

% What are the specific contributions? Start with ``Our contributions are ..."
% \input{figures/user_interface_figure}

% Context

% What we do and how that relates to existing work

% Technical overview

% Main result & contributions

% intro outline: 
% 1) highlight the growing gap between editing with generative models and traditional methods.
% 2) how our proposed solution addresses this gap to get the best of both worlds
% 3) mention the dataset issue and how videos provide a natural signal
% 4) discuss how we can warp with optical flow but that is unnatural. We segment everything, estimate affine transforms and warp. This exactly matches the user input.
% 5) highlight how by modifying the granularity of the selection and edit, the user can practically decide on the level of control in the editing process 
% 6) in our setup, we also have the original reference, and the edited image, so we use an architecture that levarages both, and encode the correspondences in the cross attention layer to facilitate preserving the input identity.

\begin{figure*}[!tbh]
% \vspace{-5mm}
\centering
% \vspace{-5mm}
\includegraphics[width=\linewidth]{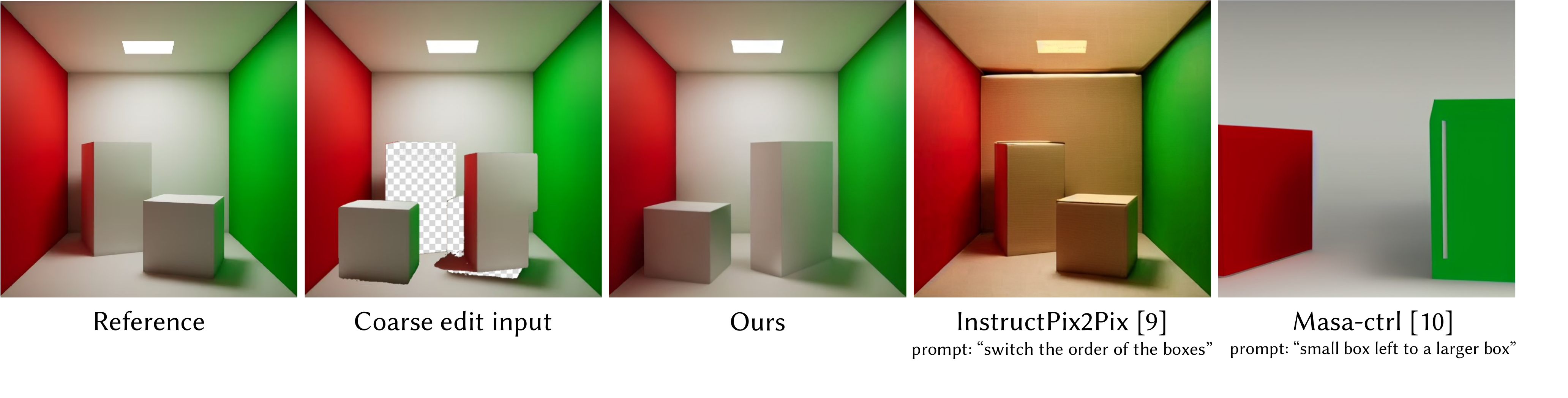}
% \mpage{0.24}{\frame{\includegraphics[width=1.0\linewidth, trim=0 0 0 0, clip]{figure_imgs/ablation_examples/all_orange_og.png}}}\hfill
% \mpage{0.24}{\frame{\includegraphics[width=1.0\linewidth, trim=0 0 0 0, clip]{figure_imgs/ablation_examples/all_orange__edit__001.png}}}\hfill
% \mpage{0.24}{\frame{\includegraphics[width=1.0\linewidth, trim=0 0 0 0, clip]{figure_imgs/ablation_examples/all_orange__edit__001_pure_noise.png}}}\hfill
% \mpage{0.24}{\frame{\includegraphics[width=1.0\linewidth, trim=0 0 0 0, clip]{figure_imgs/ablation_examples/all_orange__edit__001_output.png}}}\hfill
% \mpage{0.24}{Source} \hfill \mpage{0.24}{Edit} \hfill
% \mpage{0.24}{Output (pure noise init.)} \hfill \mpage{0.24}{Output (noisy warped init.)}
% \vspace{-5mm}
\caption{\label{fig:text_comparison}
\textbf{Comparison with text based control.} Our method directly takes a coarse user edit and makes it photorealistic. Our editing is both easy and precise, and our model can harmonize the global illumination appropriately. Text-based editing methods \cite{brooks2023instructpix2pix, cao_2023_masactrl} on the other hand, are not able to perform such edits, resulting in global appearance changes \cite{brooks2023instructpix2pix} or unrealistic image \cite{cao_2023_masactrl}.
}
% \vspace{-5mm}
\end{figure*}

% Context and state of the art
%
Image editing is a labor-intensive process.
Although humans can quickly and easily rearrange parts of an image to compose a new one, simple edits can easily look unrealistic, e.g., when the scene lighting and physical interactions between objects become inconsistent. 
Fixing these issues manually to make the edit plausible requires professional skills and careful modifications, sometimes down to the pixel level.
The success of recent generative models~\cite{rombach2021latentdiffusion,esser2021taming,ho2020denoising,dhariwal2021diffusion} paves the way for a new generation of automated tools that increase the realism of image edits while requiring much sparser user inputs~\cite{andonian2021paint,couairon2022diffedit,kim2022diffusionclip,sarukkai2024collage}.
Generative methods providing explicit spatial keypoints control have been proposed but are either limited to certain domains~\cite{pan2023draggan} or modest changes~\cite{shi2023dragdiffusion}.
State-of-the-art approaches, however, regenerate pixels based on a user-specified text prompt and a mask of the region to influence~\cite{xie2023smartbrush,wang2023imagen,brooks2023instructpix2pix,cao_2023_masactrl}.
This interface is not always natural.
In particular, it does not allow spatial transformations of the existing scene content, as we show in Figure ~\ref{fig:text_comparison}, and object identities are often not fully preserved by the re-synthesis step~\cite{chen2023anydoor, song2023objectstitch}.
%
% Key idea / high-level overview
%

In this paper, we propose a new approach to image editing that offers the controls of conventional editing methods and the realism of the modern generative model (Figure ~\ref{fig:teaser}).
Our method uses human inputs where it shines: users can segment the image and rearrange its parts manually in a ``cut-and-transform'' approach, e.g., using simple 2D transforms, duplication, or deletion to construct their desired layout, just like a collage~\cite{sarukkai2024collage}. 
We call our collage-like editing interface the \textit{Collage Transform}.
We then train a diffusion model to take care of the hard work of making the edit photorealistic.
Our model ``projects'' the coarsely edited image onto the natural image manifold, fixing up all the low-level image cues that violate its image prior, such as tweaking poses, blending object boundaries, harmonizing colors, adding cast shadows, reflections and other second-order interactions between the object and the environment.
%
% Dataset and training overview.
%

%although we start from a pre-trained diffusion model~\cite{rombach2021latentdiffusion},
Crucially, we explicitly fine-tune a latent diffusion model \cite{rombach2021latentdiffusion} so its output deviates as little as possible from the user's specifications and the appearance of the original objects in the scene.
This is essential for photographers, as they spend significant effort capturing their images and would like to retain the content identity as much as possible. 
When editing an image, there is a subtle balance between being faithful to the original image and harmonizing the edited image to preserve realism. 
This is the regime that our work focuses on.
Our insight is that videos provide a rich signal of how an edited photo's appearance should change to preserve photorealism.
From videos, we can learn how objects' appearances change in the real world as they deform and move under changing light. 
Camera motion and disocclusions give us priors about what hides behind other objects and how the same object looks under changing perspectives.

To exploit these cues, we build a paired image dataset from a large-scale video corpus.
Each pair corresponds to two frames sampled from the same video: source and target frames.
We then automatically segment~\cite{kirillov2023segment}, and transform objects in the source frame to match the pose of the corresponding objects in the target frame, using two motion models based on optical flow, designed to simulate the coarse edits a user would make using our Collage Transform interface.
Since the images are now roughly aligned, we can train our model to convert the coarsely edited image into the ground truth target frame in an image-to-image~\cite{saharia2022palette,isola2017image} fashion.
This alignment procedure encourages the model to follow the user-specified layout at test time closely.
Additionally, our model is carefully designed to transfer fine details from the reference source frame to preserve the identity and appearance of objects in the scene.

Our approach can produce plausible and realistic results from real user edits, and effectively projects coarse user edits into photorealistic images, confirming our insights on the advantages of using video data and a carefully designed motion model.
Compared to the state-of-the-art, we show our outputs are preferred 89\% of the time in a user study.

In short, our contributions are as follows:
\begin{itemize}
    \item the Collage Transform, a natural interface for image editing that allows users to select and alter any part of an input image using simple transforms and that automatically turns the resulting edit into a realistic image,
    \item a new paired data generation approach to supervise the conversion from coarse edits to real images, which extracts pairs of video frames and aligns the input with the ground truth frame using simple motion models, 
    % \item a diffusion model design that uses the warped image to guide the layout of the generated image and transfers details from the source reference frame to match the ground truth. 
    \item a conditioning procedure that uses: 1. the warped image to guide layout in the diffusion generator, and 2. features from a second diffusion model to transfer fine image details and preserve object identity.
    \item a comprehensive analysis on the model's generalization to diverse editing tasks, like spatial editing, colorization, 3D transformation, NS perspective warping.
\end{itemize}

\section{Related Work}
\label{sec:related}

% Relationship examples:
% Similar:
% - Our work also adopt X ...
% - We address similar challenges.
% Different:
% - Our work differs in X ...
% - In contrast, we tackle ...

\myparagraph{Classical image editing.}
Classical image editing techniques offer various types of user controls to achieve diverse objectives. For instance, image retargeting aims to alter an image's size while preserving its key features and content \cite{avidan2007carving,rubinstein2008improved, wang2008optimized, simakov2008summarizing}. In contrast, image reshuffling rearranges an image's content based on user-provided rough layouts and imprecise mattes \cite{simakov2008summarizing, cho2008patch, barnes2009patchmatch}. Image harmonization integrates objects from different images, adjusting their low-level statistics for a seamless blend \cite{jia2006dragdrop, sunkavalli2010ms_harmonization}. 
% Interactive photomontage, on the other hand, merges various images taken from the same perspective \cite{agarwala2004interactive}. 
A common thread in these classical image editing applications is the crucial role of user interaction, which provides the necessary control for users to realize their vision. Our method aligns with this approach, allowing users to reconfigure a photograph based on their preliminary edits. 
% It then employs a generative model to refine these edits, ensuring they are photorealistic and seamlessly integrated.
% The patch transform \cite{cho2009patch} allows the user to spatially recompose the image by moving patches of the image around, while maintaining consistent boundaries between the patches (similar to our method in spirit, but they use classical optimization. As a result, the method is somewhat limited as they wouldn't relight, add or remove shadows, and higher level -- squared patches rather than arbitrary segments)

% classical image harmonization methods focus on adjusting low-level statistics \cite{jia2006dragdrop, sunkavalli2010ms_harmonization}, and is focused on harmonizing objects from another image. Our use case is spatially recomposing a single image. Similarly, interactive photomontage \cite{agarwala2004interactive}
% combines different images taking from the same pose.

%     \item collage editing (eli mentioned an old classical paper a while back can't recall it)
% \end{itemize}

\myparagraph{Controllable image generation.} The rapid advancement in photorealistic image generation has inspired researchers to adapt generative models for image editing tasks. Early efforts focused on high-level edits, like altering age or style, by manipulating latent space of Generative Adversarial Networks (GANs) \cite{abdal2019image2stylegan, abdal2020image2stylegan++, chai2021using}. In a vein similar to our work, Generative Visual Manipulation \cite{Zhu2016GenerativeVM} involves projecting user-edited images onto the natural image manifold as approximated by a pre-trained GAN. The recent introduction of CLIP embeddings \cite{radford2021learning} has further propelled image editing capabilities, particularly through text prompts \cite{avrahami2022blended, crowson2022vqgan, gal2022stylegan, kim2022diffusionclip, brooks2023instructpix2pix, hertz2022prompt, mokady2023null}. 
% Among these methods, those that utilizing the robust generative prior of diffusion models have achieved impressive photorealistic results. 
% Moreover, approaches like universal guidance \cite{bansal2023universal} and readout guidance \cite{luo2023readoutguidance} influence the iterative denoising process of diffusion models for guided image generation. 
DragGAN \cite{pan2023draggan} introduces fine control in image editing by using key-handles to dictate object movement, and follow-up works extend the drag-control idea to diffusion models \cite{shi2023dragdiffusion, mou2023dragondiffusion, luo2023readoutguidance}. 
Image Sculpting \cite{yenphraphai2024sculpting} takes a different approach by directly reposing the reconstructed 3D model of an object and re-rendering it, providing high level of control, but time consuming editing process unlike our Collage Transform interface that is designed to increase editing efficiency. 
CollageDiffusion \cite{sarukkai2024collage} guides text-to-image generation by using a collage as additional input. However, while CollageDiffusion focuses on controlling the generation of an image from scratch, we focus on using collage-like transformation to edit a reference image, and focus on preserving its identity.

% Generative models have indeed shown significant effectiveness in image compositing tasks \cite{yang2023paint, chen2023anydoor, song2023objectstitch}. Paint by Example specializes in inpainting an image using content from a reference image. ObjectStitch \cite{song2023objectstitch} introduces a self-supervised training approach for a conditional diffusion model tailored for object insertion, offering more control than traditional reference-based inpainting methods. Similarly, AnyDoor \cite{chen2023anydoor} trains its conditional model using video frames. However, both Paint by Example and ObjectStitch \cite{song2023objectstitch} provide only coarse user control, typically through a bounding box to determine the location for object insertion, and they struggle with preserving the identity of the objects. In contrast, our method offers much greater flexibility for editing objects within an image, such as altering their pose, while also maintaining a higher level of identity preservation.

\myparagraph{Reference-based editing with generative models.}
To extend controllable image generation into editing real (non-generated images), one can invert the image back to noise \cite{song2020denoising}, and then guide the iterative denoising process to control the image generation\cite{bansal2023universal, meng2022sdedit, cao_2023_masactrl}. However, naively guiding the model without any grounding can lead to a loss in image identity. Prior work \cite{yang2023paint, epstein2023selfguidance, chen2023anydoor} preserves the image identity through a pretrained feature extractor like CLIP \cite{radford2021learning} or DINO \cite{oquab2023dinov2}, using a Control-Net like feature-injection \cite{zhang2023adding, chen2023anydoor}, a dual-network approach \cite{cao_2023_masactrl, hu2023animateanyone}, or a combination of those approaches \cite{chen2023anydoor, xu2023magicanimate}. We adopt the dual-network approach, as it allows us to fully fine-tune the model and taylor it to our photorealistic editing task using our video-based dataset. AnyDoor~\cite{chen2023anydoor} similarly uses video frames during training, but their focus is to recompose individual objects into the scene. On the other hand, we use video data to recompose the \textit{entire scene} and use motion models designed for a convenient photo editing interface. Closest to our work is MotionGuidance \cite{geng2024motion} that uses optical flow to guide editing the reference frame with diffusion guidance \cite{bansal2023universal} for a highly user-controllable edit. However, dense optical flow is difficult to manually provide for a user, unlike simple cut-and-transform edits in our Collage Transform. Furthermore, they rely on a prohibitively time-consuming guidance that take as long as 70 minutes for a single sample. On the other hand, our approach takes less than 5 seconds to fix up the user edit, allowing for interactive editing process.

\begin{figure*}[!htb]
\centering
\vspace{-5mm}
\includegraphics[width=\linewidth]{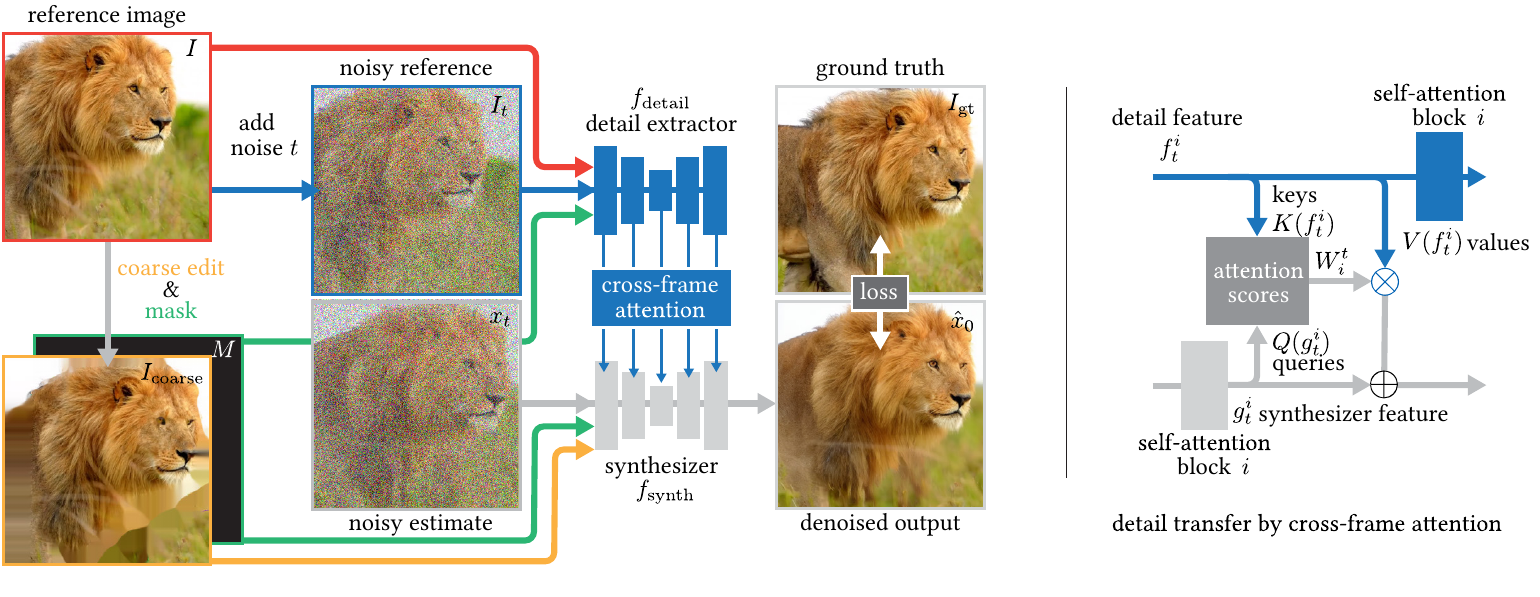}
\vspace{-5mm}
\caption{\label{fig:overview_figure}
{\bf Overview.}
Our pipeline (left panel) uses \new{two parallel models}, a detail extractor (top) and a synthesizer (bottom), to generate a realistic image from a coarse user edit and a mask recording missing regions caused by the edit.
The detail extractor processes the reference image, a noisy version of the reference and the mask, to produce a set of features that guide the synthesis and allow us to preserve the object appearance and fine details from the reference image.
The synthesizer generates the output conditioned on the mask and coarse edit. The features from the detail extractor are injected via cross-attention at multiple stages in the synthesizer, in order to transfer details from the input.
Both models are finetuned on our paired dataset.
The right panel shows a detailed view of our cross-attention detail transfer operator.
Photos sourced from \copyright Adobe Stock.
% This figure illustrates our training pipeline.
% %
% During inference (not shown), we initialize $x_T$ by adding noise to the coarse edit $I_\text{coarse}$ as described in Section~\ref{sec:conditional-diffusion}.
%
% \hadi{R1 some of the coloring being confusing}
}
% \vspace{-5mm}
\end{figure*}

\section{Method}
\label{sec:method}

% \mg{some of that can become intro material}
% Realistic image editing remains a challenging problem.
% %
% Specifying coarse image modifications using rough selections and simple transformations is quick and straightforward for a human user, but rarely produces realistic results.
% %
% Altering a photograph in a way that preserves realism in layout, lighting and general appearance often
% requires tedious editing down to the pixel level, which is time consuming.
% %
% Gen models produce good-looking harmonized results, but difficult to control, limited to n identity preservation, only loosely follows use controls

We aim to enable an image editing workflow in which users can select objects in a photograph, duplicate, delete or rearrange them using simple 2D transforms to produce a realistic new image (\S~\ref{sec:ui}).
We leverage image priors from pretrained diffusion models to project the coarsely edited image onto the natural image manifold, so the user can focus on specifying high-level changes without worrying about making their edits plausible (\S~\ref{sec:diffusion-projection}).
Existing diffusion models can produce impressive results but often do so at the expense of control and adherence to the user input~\cite{meng2022sdedit}.
In particular, they tend to ``forget'' the identity and appearance of the edited object~\cite{yang2023paint}, and often only loosely conform to the user-specified pose~\cite{chen2023anydoor}.
Our method addresses these issues using two mechanisms.
First, our synthesis pipeline is a conditional diffusion model (\S~\ref{sec:conditional-diffusion}) that follows the coarse layout defined by the user, and transfers fine details from the reference input image (\S~\ref{sec:detail-transfer}) to best preserve the original image content.
Second, we construct a supervised dataset exploiting object motion from videos to finetune the pretrained model to explicitly encourage content preservation and faithfulness to the input edit (\S~\ref{sec:video-motion}).

\subsection{Specifying coarse structure with simple transforms}\label{sec:ui}

Starting from an image $I\in\mathbb{R}^{3hw}$, $h=w=512$, we run an automatic segmentation algorithm~\cite{kirillov2023segment} to split the image into non-overlapping semantic object segments.
The user can edit this image by applying 2D transformations to the individual segments (e.g., translation, scaling, rotation, mirroring).
Segments can also be duplicated or deleted.
Figure~\ref{fig:teaser} illustrates this workflow.
We keep track of holes caused by disocclusions when moving the segment in a binary mask $M\in\{0,1\}^{hw}$, and inpaint them using a simple algorithm~\cite{Bertalmo2001NavierstokesFD}.
We denote the resulting, coarsely edited image by $I_\text{coarse}\in\mathbb{R}^{3hw}$.

We operate in an intermediate latent space for efficiency, but our approach also applies to pixel-space diffusion.
With a slight abuse of notation, in the rest of the paper $I,I_\text{coarse}\in\mathbb{R}^{3hw}$, with $h=w=64$ refer to the input and coarse edit after encoding with the latent encoder from Stable Diffusion~\cite{rombach2021latentdiffusion}, and $M$ the mask downsampled to the corresponding size using nearest neighbor interpolation.
The latent triplet $(I,I_\text{coarse}, M)$ forms the input to our algorithm.

\subsection{From coarse edits to realistic images using diffusion}\label{sec:diffusion-projection}

We want to generate a realistic image that
\begin{inparaenum}[(1)]
\item follows the large-scale structure defined by the coarse user edit, and
\item preserves the fine details and low-level object appearance from the unedited image, filling in the missing regions.
\end{inparaenum}
Our pipeline, illustrated in Figure~\ref{fig:overview_figure}, uses 2 parallel models.

The first, which we call \emph{synthesizer} $f_\text{synth}$, generates our final output image.
%
% It is conditioned on the  user's coarse edit, $I_\text{coarse}$ and the edit mask $M$.
%
The second model, which we name \emph{detail extractor} $f_\text{detail}$, transfers fine-grained details from the unedited reference image $I$ to our synthesized output during the diffusion process.
It modulates the synthesizer by cross-attention at each diffusion step, an approach similar to Masa-Ctrl \cite{cao_2023_masactrl} and AnimateAnyone \cite{hu2023animateanyone}.
Both models are initialized from a pretrained Stable Diffusion v1.4 model~\cite{rombach2021latentdiffusion}, and finetuned on our paired dataset (\S~\ref{sec:video-motion}).
Since we have a detailed reference image $I$ to guide the synthesis, we do not need the coarse semantic guidance provided by CLIP, so we remove the CLIP cross-attention from the model.
%
% \mg{this prompt an ablation: what happens if we just pass the ref by concatenation}.
%
% Since the CLIP cross-attention is limited to providing semantic information, we replace the CLIP cross-attention layer with a cross-attention layer that attends to the reference image (query computed from the edited image -- key and value from the reference image features -- need to write this cleanly).

Let $T\in\mathbb{N}^*$ be the number of sampling steps, and $\alpha_0,\ldots,\alpha_T\in\mathbb{R}^{+}$ be the alphas of the diffusion noise schedule~\cite{ho2020denoising}.
Starting from an image $x_0\in\mathbb{R}^{3hw}$, the forward diffusion process progressively adds Gaussian noise, yielding a sequence of increasingly noisy iterates:
\begin{equation}
x_t \sim \mathcal{N}\left(\sqrt{\alpha_t}x_{t-1}; (1-\alpha_t)\mathbf{I}\right).
\end{equation}
The base diffusion model $f$ is trained to reverse this diffusion process
% , $x_{t-1} = f(x_t; t)$, 
and synthesize an image iteratively, starting from pure noise $x_T\sim\mathcal{N}(0, I)$.
The synthesizer and detail extractor in our approach make a few modifications to this base model, which we describe next.

\subsection{Extracting details from the reference image}\label{sec:detail-transfer}
% \subsection{Extracting details from the reference image transfer from the reference using cross-attention}\label{sec:detail-transfer}
%
During inference, at each time step $t$, we start by extracting a set of features $F_t$ from the reference image using $f_\text{detail}$ (Figure~\ref{fig:overview_figure}, top). 
These features will guide the synthesis model and help preserve realistic image details and object identity.
Since we use a pretrained diffusion model as a feature extractor, we start by adding noise to the reference unedited image:
\begin{equation}
I_t = \sqrt{\bar\alpha_t} I + (1-\bar\alpha_t)\epsilon,
\end{equation}
with $\epsilon\sim\mathcal{N}(0, \mathbf{I})$, $\bar\alpha_t = \prod_{s=1}^{t}\alpha_s$.
We extract the feature tensors immediately before each of the $n=11$ self-attention blocks in the model:
\begin{equation}
F_t := [f_t^1,\ldots,f_t^n] = f_\text{detail}([I_t, I, M]; t),
% [\phi^0_t,\ldots,\phi^n_t] := f_\text{detail}(I_t, I; t),
\end{equation}
where $[\cdot]$ denotes concatenation along the channel dimension.
Our feature extractor also takes as input the clean reference image since it is always available for detail transfer and mask, so the model knows which regions need inpainting.
Since the pretrained model only takes $I$ as an input, we modify the first layer at initialization by padding its weight with zeros to accept the additional channel inputs.
Using a noisy version of the reference ensures the extracted features are comparable to those in the cross-attention operators of the synthesis model. 
%
% Our synthesis model $f_\text{synth}$ 

% \begin{equation}
%     \
% \end{equation}

% \begin{equation}
% z_{t+1} = f(z_t;z_\text{coarse}, z)
% \end{equation}

% Our input consists of the edited image, and original reference, and a mask that specifies the missing regions from the edited image (as flow warping and collage editing creates holes in where the piece was).

\subsection{Image synthesis by detail transfer to the coarse edit}\label{sec:conditional-diffusion}

The synthesizer $f_\text{synth}$ generates the final image, conditioned on the detail features $F_t$.
Unlike standard diffusion sampling, we do not start from pure Gaussian noise.
Instead, inspired by SDEDit ~\cite{meng2022sdedit}, we start from an extremely noisy version of the coarsely edited image:
\begin{equation}
    x_{T-1} = \sqrt{\bar\alpha_{T-1}} I_\text{coarse} + (1-\bar\alpha_{T-1})\epsilon,
\end{equation}
so we effectively bypass the first denoising step with adding noise to the edit rather than pure noise. This initialization circumvents a commonly observed issue where diffusion models struggle to generate images whose mean and variance deviate from the normal distribution. This is particularly important in our setup as the user input can have arbitrary color distribution, and we need the model to match the user input.
\begin{figure*}[!bth]
\centering
\includegraphics[width=\linewidth]{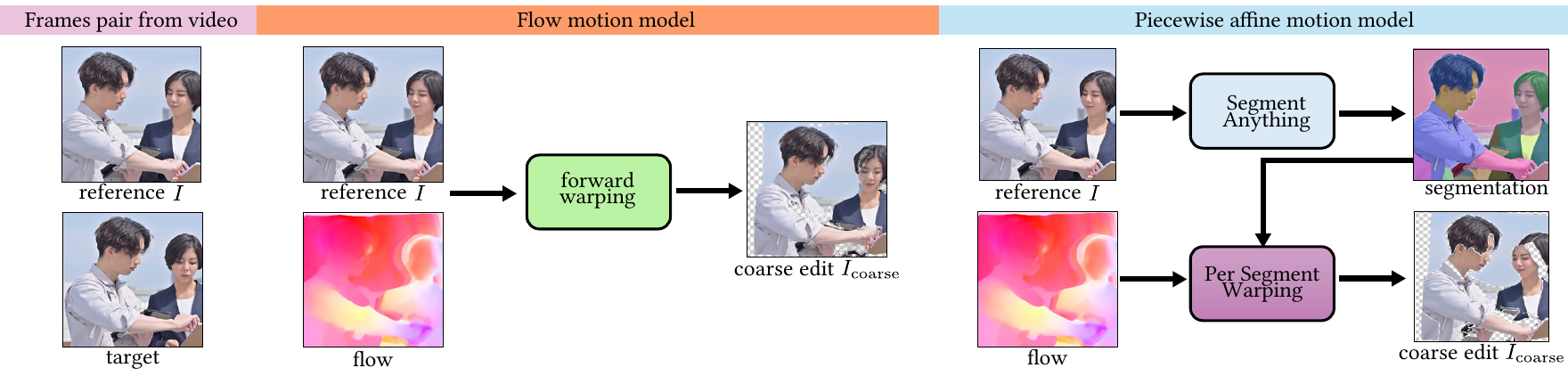}
% \vspace{-7mm}
\caption{\label{fig:motion_models} \new{
\textbf{Dataset synthesis.} Our dataset synthesis pipeline starts by sampling two frames from a video. We set one frame to be the reference, which we warp by one of our motion models, and set the other frame to be the target that we warp the reference towards.
To generate aligned training pairs, we use 2 motion model to warp the reference frame towards the ground truth (target frame).
The first model uses optical flow (left).
It provides the most accurate alignment but does not correspond to what the user would provide during inference.
This motion model encourages adherence of our model's output to the layout specified using the coarse edit.
To generate training pairs closer to the collage-like user inputs, we use a second motion model (right).
For this, we segment everything in the image~\cite{kirillov2023segment} and apply similarity transforms to each segment, estimated from the flow within the segment.
Figure~\ref{fig:motion_model_ablation} analyses the impact of these motion models on the final result.
% \hadi{I vote change this to be one row of frames (1x8 rather than 2x4)}
%
Photos sourced from \copyright AdobeStock.
}
}
% \vspace{-5mm}
\end{figure*}

\begin{figure}[!tbh]
\centering
\includegraphics[width=\linewidth]{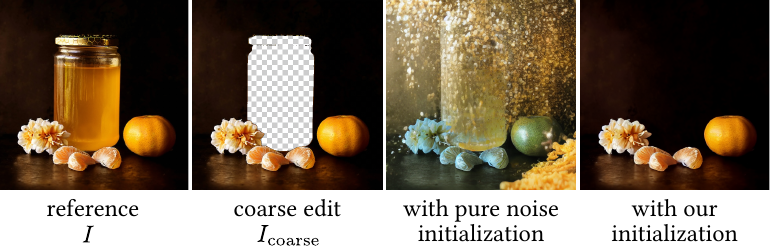}
% \mpage{0.24}{\frame{\includegraphics[width=1.0\linewidth, trim=0 0 0 0, clip]{figure_imgs/ablation_examples/all_orange_og.png}}}\hfill
% \mpage{0.24}{\frame{\includegraphics[width=1.0\linewidth, trim=0 0 0 0, clip]{figure_imgs/ablation_examples/all_orange__edit__001.png}}}\hfill
% \mpage{0.24}{\frame{\includegraphics[width=1.0\linewidth, trim=0 0 0 0, clip]{figure_imgs/ablation_examples/all_orange__edit__001_pure_noise.png}}}\hfill
% \mpage{0.24}{\frame{\includegraphics[width=1.0\linewidth, trim=0 0 0 0, clip]{figure_imgs/ablation_examples/all_orange__edit__001_output.png}}}\hfill
% \mpage{0.24}{Source} \hfill \mpage{0.24}{Edit} \hfill
% \mpage{0.24}{Output (pure noise init.)} \hfill \mpage{0.24}{Output (noisy warped init.)}
\caption{\label{fig:latent_init}
\textbf{Effects of Latent Initialization.} Starting from pure noise, as is standard practice, the model struggles to generate images with deep blacks and synthesizes nonsensical content to keep the image's mean and standard deviation close to the starting Gaussian noise. This is a known issue with current diffusion models~\cite{guttenberg2023offset, lin2024snr}.
Instead, during inference, we initialize the latent to the warped image with a very large amount of Gaussian noise before running the diffusion.
This simple change makes a drastic difference and lets the model preserve the image content.
Photo sourced from \copyright Unsplash.
}
\end{figure}

This has been shown to stem from a domain gap between training and sampling~\cite{guttenberg2023offset,lin2024snr}: the model never sees pure noise during training, but a sample from the normal distribution is the starting point for inference. Our latent initialization addresses this issue by directly bridging the gap between training and inference. In Fig.~\ref{fig:latent_init} we show how initializing with pure noise leads to a low contrast image, while our initialization allows the model to preserve the input color range well.
For subsequent steps during inference, we update the current image estimate $x_t$ at each time step $t$, using the following update rule:
\begin{equation}
x_{t-1} = f_\text{synth}([x_t, I_\text{coarse}, M];t, F_t).
% x_{t+1} := f_\text{synth}(x_t, I_\coarse, M; t, [\phi^0_t,\ldots,\phi^n_t]),
\end{equation}
We provide the mask and coarse edit as conditions by simple concatenation, but because we need to extract fine details from the reference, we found passing the reference information by cross-attention with the features $F_t$ provided richer information. 
Again, we extend the weight tensor of the first convolution layer with zeros to accommodate the additional input channels.

% \hadi{I'm confused by the last two sentences. seems like you intended to say something else but then mentioned initializing weights}

% to transfer details from the reference frame $I$.

% \paragraph{Sampling initialization}

% During i
% Prior works showed 
%
% during training, the model processes a noisy version $x_t$ of a ground truth image $x_0$, but never start from 

%
% To address the issue, prior works suggested modified training schedules where the model takes a pure noise input with zero signal-to-noise ratio (SNR), and proposed different variations to the training process. 
%
% However, for our problem setting we can directly address the domain gap by initializing the latent to denoise by adding large amount of noise to the warped latent. 
%
% (For a sampling schedule of 1000 steps, instead of starting from step 0 with pure noise, we just start from step 1 so there is incredibly large amount of noise and very small signal is introduced -- need to explain this better). 
%
% We condition the synthesis model on the user-specified image layout, represented by the coarse edit $I_\text{coarse}$ and mask $M$.
% trained to reverse this diffusion process, starting from $x_T$ to synthesize an image.

\paragraph{Detail transfer via cross-attention}
We use the intermediate features $F_t=[f_t^1,\ldots,f_t^n]$, extracted \emph{before} the detail extractor's self-attention layers to transfer fine image details from the reference image to our synthesis network by cross-attention with features $[g_t^1,\ldots,g_t^n]$ extracted \emph{after} the corresponding self-attention layers in the synthesis model. % (both models have the same architecture).
See the right panel of Fig.~\ref{fig:overview_figure} for an illustration, where $Q$, $K$, $V$ are linear projection layers to compute the query, key, and value vectors, respectively, and $W_i^t$ is the matrix of attention scores for layer $i$, at time step $t$.
The feature tensors $g_t^i,f_t^i$ are 2D matrices whose dimensions are the number of tokens and feature channels, which depend on the layer index $i$.

% , we compute the output features as
%
% \begin{equation}
%     g_i = \sum_{j=1}^n \text{softmax}\left(\frac{q(g_i)\cdot k(f_j)}{\sqrt{d}}\right) v(f_j)
%     % F_\text{synth} = \left(\frac{Q^TK}{\sqrt{d}} V\right)
% \end{equation}
% %
% \mg{not sure whether we need an explicit equation for the cross attention, since we add a layer, but maybe we need to explain the queries come from the synth model and the keys from the detail model}

% and $F_t$.
% is a set of intermediate features extracted from the reference image that interact with the synthesis model via cross-attention, which we describe next.

\subsection{Training with paired supervision from video data}\label{sec:video-motion}

We \new{train our model} on a new dataset obtained by extracting image pairs from videos to reconstruct a ground truth frame given an input frame and a coarse edit automatically generated from it.
%
% \mg{r5: not clear how to finetune models that accept 3 input images, from the pretrained diffusion model}
% %
% \mg{R2: explain where data comes from. "proprietary dataset"?}
%
Our insight is that motion provides useful information for the model to learn how objects change and deform.
Videos let us observe the same object interact with diverse backgrounds, lights, and surfaces. 
For example, skin wrinkles as a person flexes their arm, their clothes crease in complex ways as they walk, and the grass under their feet reacts to each step.
Even camera motion yields disocclusion cues and multiple observations of the same scene from various angles.

% To ensure our pipeline generates plausible images whose content and details are consistent with the reference input, 
%
% Training our pipeline in a supervised manner lets our synthesizer learn to transform coarsely edited image into realistic images exhibiting these complex behaviors, which would be extremely difficult to produce by hand. 

Concretely, each training sample is a tuple $(I, I_\text{gt}, I_\text{coarse}, M)$, where $I$ and $I_\text{gt}$ are the input and ground-truth frames, respectively, extracted from the video with a time interval sampled uniformly at random from $\{1,\ldots,10\}$ seconds between them. However, if the computed flow between the two frames was too large (at least 10 percent of the image has a flow magnitude of 350 pixels), we resample another pair. This is to ensure that the warping produces reasonable outputs.
We construct the coarse edit $I_\text{coarse}$ and corresponding mask $M$ using an automated procedure that warps $I$ to approximately match $I_\text{gt}$, in a way that mimics our Collage Transform interface.
For this, we use one of 2 possible editing models: a flow-based model and a piecewise affine motion model (Fig~\ref{fig:motion_models}).

\paragraph{Flow-based editing model}
We compute the optical flow using RAFT-Large~\cite{teed2020raft} for each consecutive pair of frames between $I$ and $I_\text{gt}$ and compose the flow vectors by backward warping the flow to obtain the flow between the two frames.
We then forward warp $I$ using softmax-splatting \cite{niklaus2020softsplatting}, to obtain $I_\text{coarse}$, which roughly aligns with the ground truth frame.
The forward warping process creates holes in the image. 
We record these holes in the mask $M$.
Our model needs to learn to inpaint these regions and those we have no correspondence (e.g., an object appearing in the frame). 
Using flow-based warping helps the model learn to preserve the identity of the input, rather than always hallucinating new poses and content.

\paragraph{Piecewise affine editing model}
Optical flow warping can sometimes match the ground truth too closely.
As we discuss in Section~\ref{sec:result} and Figure~\ref{fig:motion_model_ablation}, training the flow-based editing model only can limit the diversity of our outputs, leading to images that do not deviate much from the coarse edit.
Flow-warping is also reasonably distinct from our expected test-time user inputs (\S~\ref{sec:ui}).
Our second editing model addresses these issues by transforming the reference frame as a \textit{collage}.
We compute a depth map for the image using MiDaS \cite{Ranftl2020midas, Ranftl2021midas} and automatically segment the image using SegmentAnything~\cite{kirillov2023segment}.

We then transform each segment using the affine transformation that best matches the optical flow for this segment,
compositing them back to front according to each segment's average depth.
For the image regions that are not segmented, we use the optical flow warping scheme described above.
Due to the coarser alignment, the model learns how different parts of the scene interact in a realistic setting, like associating objects and their shadows and reflections.

% Get the segments with SAM. We transform each piece in the order based on their estimated relative-depth (to perform something like the painter's algorithm), and estimate the depth using 

We use a dataset consisting of 12 million 5-10 second clips of stock videos, and we filter out keywords that indicate static scenes or synthetic/animated videos, as we are only interested in photo-realistic videos and also highly dynamic scenes where the motion is too large (like car racing). For each valid clip, we sample one pair and compute the warping using both motion models. After filtering for desired motion, we use 2.5 million clips, creating a dataset consists of 2.5 million samples for each motion model, making a total of 5 million training pairs.

\subsection{Implementation details}

We finetune both models jointly for 120,000 steps with a batch size of 32, using Adam \cite{Kingma2014AdamAM}, with a learning rate of $1\times 10^{-5}$ on 8 NVIDIA A100 GPUs, which takes approximately 48 hours. Note that this is considerably more efficient than recent compositing work \cite{yang2023paint} that uses 64 NVIDIA V100 GPUs for 7 days. We hypothesize that the stronger input signal helps the model converge faster.
We use a linear diffusion noise schedule, with $\alpha_1=0.9999$ and $\alpha_T=0.98$, with $T=1000$. During inference, we sample using DDIM for 50 denoising steps.

\begin{figure*}[!htp]
% \begin{center}
\centering
\includegraphics[width=\linewidth]{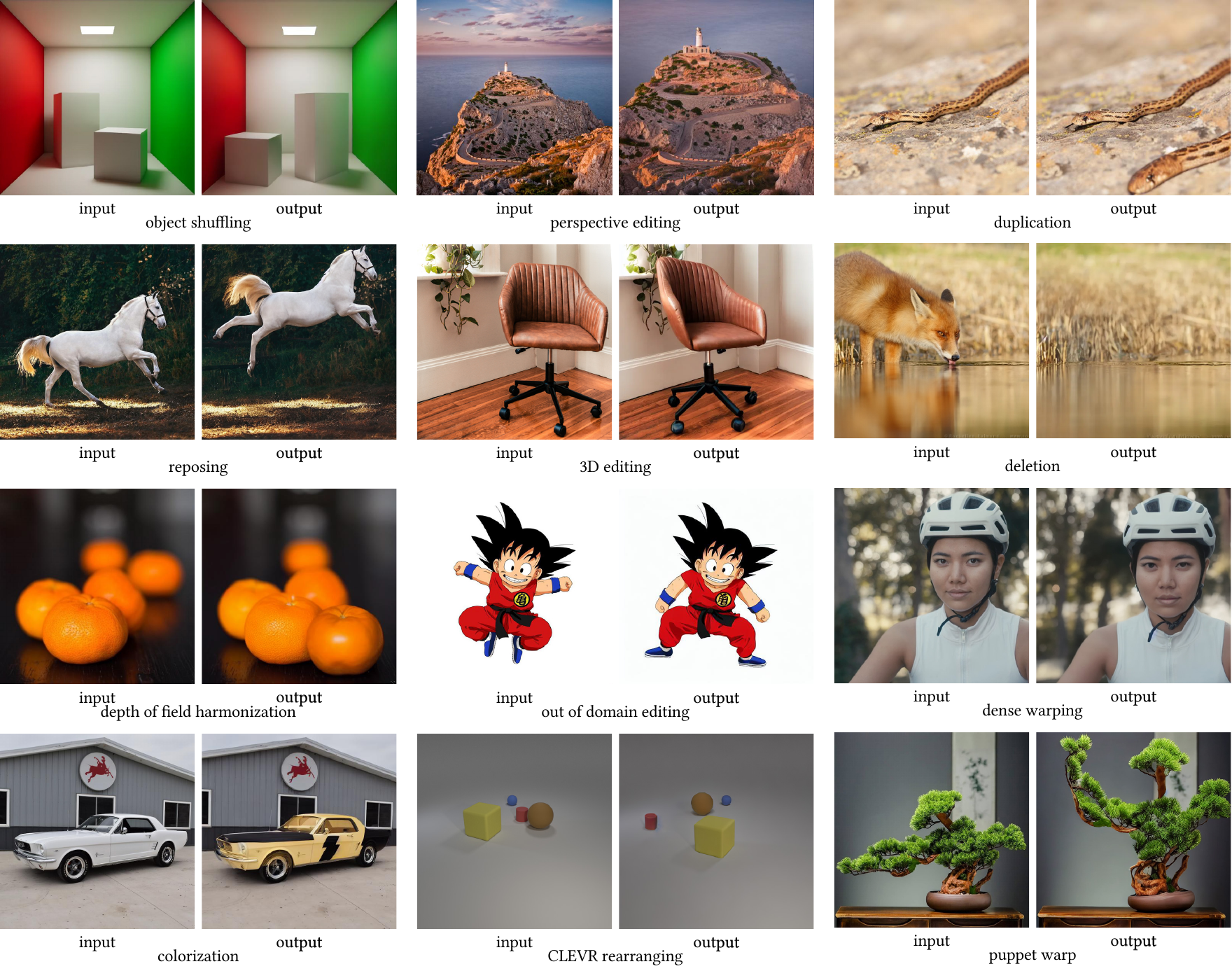}
\caption{\label{fig:diverse_applications}
\new{
\textbf{MagicFixup applications.} Our method can generalize to a large number of different applications. We just need to create a coarse edit, and MagicFixup realistically clean up the edit. Here we show a selection of tasks we used MagicFixup to achieve, like image recomposition, perspective editing, reposing, 3D transformation, and colorization. We show the coarse edits for these examples throughout the appendix. Photos sourced from \copyright Unsplash and CC licensed images.
}
}
\end{figure*}

% synthesizing realistic moving sand (row 3), fine hairs that blend convincingly with the background (row 4), 

\begin{figure*}[!htp]
% \begin{center}
\centering
\includegraphics[width=\linewidth]{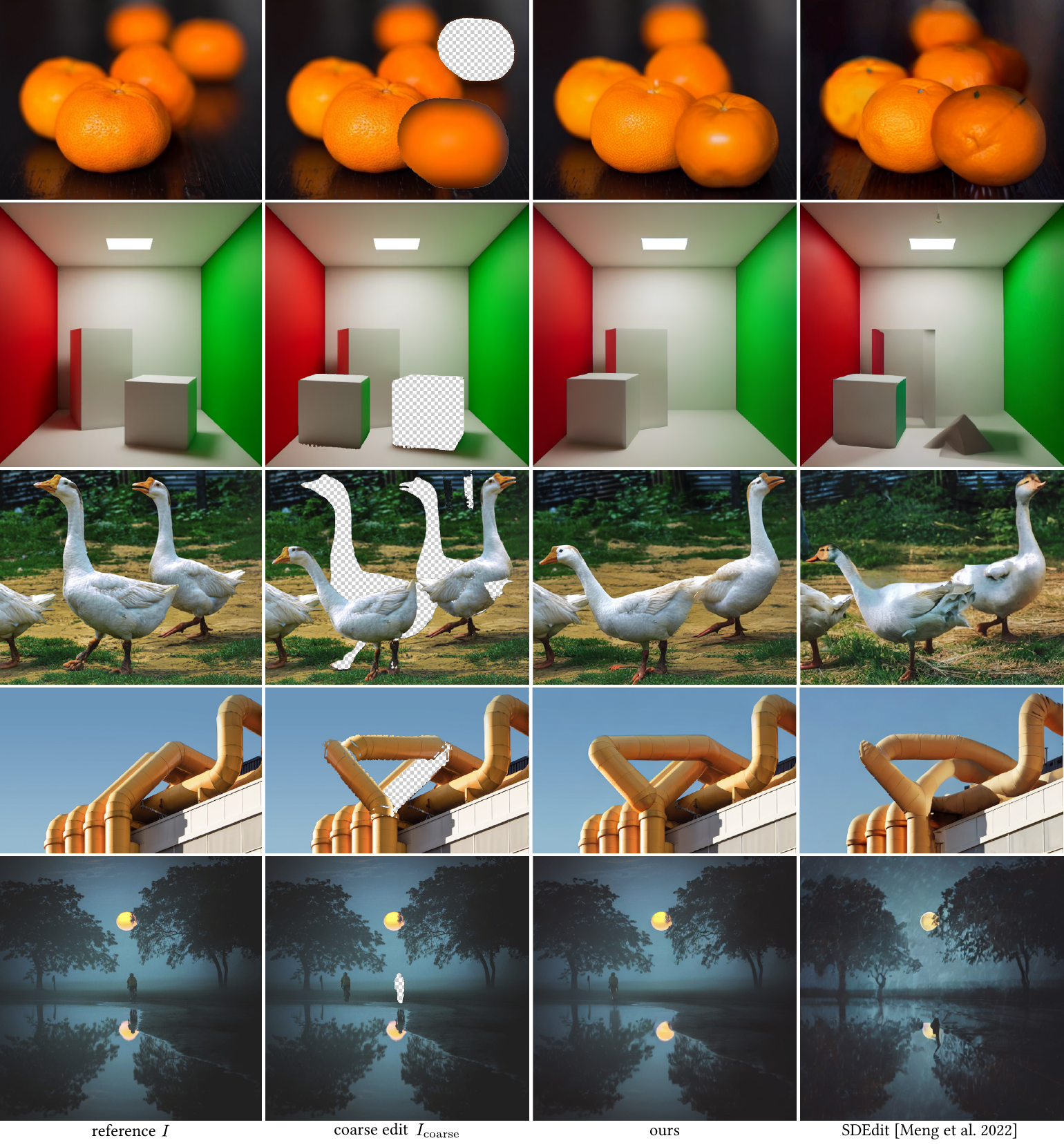}
\caption{\label{fig:applications}
\textbf{Spatial editing.} We show example of scene recompositing. Our model is capable of synthesizing compelling effects that harmonize realistically with the rest of the image such as: changing the depth of field (row 2),
adjusting the global illumination (green reflection on the cube, row 3),
and removing or adding reflections (row 6). Photos sourced from \copyright Unsplash.
}
\end{figure*}

% synthesizing realistic moving sand (row 3), fine hairs that blend convincingly with the background (row 4), 

\begin{figure*}[!hbt]
% \vspace{-3mm}
\centering
\includegraphics[width=1.0\linewidth]{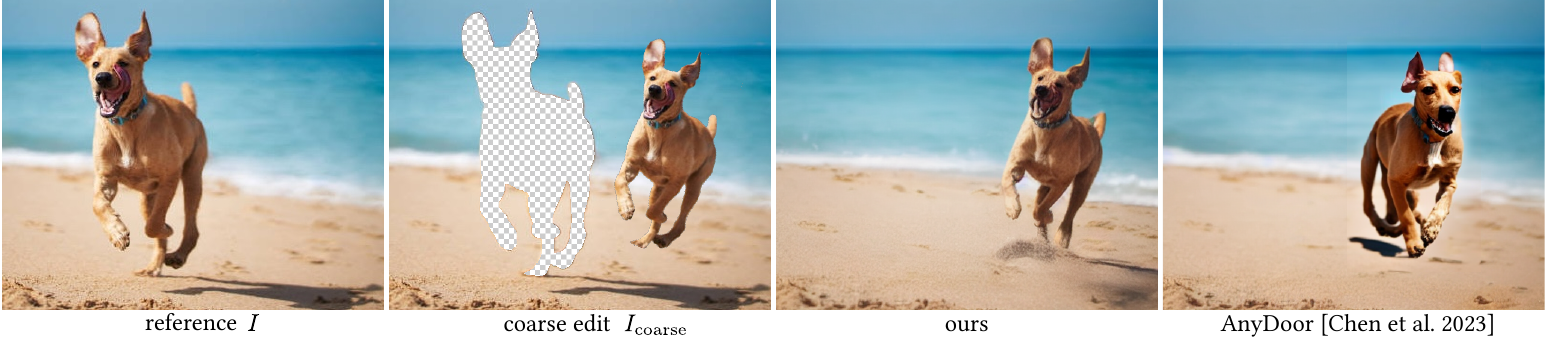}
% \mpage{0.24}{\frame{\includegraphics[width=1.0\linewidth, trim=0 0 0 0, clip]{figure_imgs/applications_images/dog_beach_og.png}}}\hfill
% \mpage{0.24}{\frame{\includegraphics[width=1.0\linewidth, trim=0 0 0 0, clip]{figure_imgs/applications_images/dog_beach__edit__009.png}}}\hfill
% \mpage{0.24}{\frame{\includegraphics[width=1.0\linewidth, trim=0 0 0 0, clip]{figure_imgs/applications_images/dog_beach__edit__009_output.png}}}\hfill
% \mpage{0.24}{\frame{\includegraphics[width=1.0\linewidth, trim=0 0 0 0, clip]{figure_imgs/anydoor_out/anydoor_dog.png}}}\hfill
% \mpage{0.24}{\frame{\includegraphics[width=1.0\linewidth, trim=0 0 0 0, clip]{figure_imgs/applications_images/snake_og.png}}}\hfill
% \mpage{0.24}{\frame{\includegraphics[width=1.0\linewidth, trim=0 0 0 0, clip]{figure_imgs/applications_images/snake__edit__003.png}}}\hfill
% \mpage{0.24}{\frame{\includegraphics[width=1.0\linewidth, trim=0 0 0 0, clip]{figure_imgs/applications_images/snake__edit__003_output.png}}}\hfill
% \mpage{0.24}{\frame{\includegraphics[width=1.0\linewidth, trim=0 0 0 0, clip]{figure_imgs/anydoor_out/anydoor_snake.png}}}\hfill
% \mpage{0.24}{Reference image} \hfill \mpage{0.24}{User edit} \hfill
% \mpage{0.24}{Our Output} \hfill \mpage{0.24}{AnyDoor \cite{chen2023anydoor}}
\vspace{-5mm}
\caption{
\textbf{Comparison to Anydoor \cite{chen2023anydoor}.} Anydoor was trained to insert objects from one image to another. We can repurpose their approach for our image editing task by using the same image as source and target.
Their approach does not preserve the dog's identity in this example. AnyDoor also does not harmonize the lighting properly (the sun direction and shadows are wrong), the image is too bright, and some blending seams are visible.
On the other hand, our output shows natural shadows and plausible contacts with the ground, adding realistic moving sand consistent with the pose. 
Photo sourced from \copyright Unsplash.
}
% \vspace{-5mm}
% Two pipelines approach. Parallel pipeline to extract features from the original image, and injects attention using correspoendences and cross-reference attention. Using DINO for feature extractions. Finetune the entire model.
\label{fig:insertion_comparison}
\end{figure*}

\section{Experimental Results}
\label{sec:result}

% high level
We evaluate our method qualitatively on a set of user edits to demonstrate real-world use cases, as well as on a held-out validation dataset created in the same way as our training set (\S~\ref{sec:video-motion}) for quantitative evaluation. \textit{In the appendix, we show additional applications of the model on editing tasks beyond spatial recomposition, like colorization, perspective editing, and 3D transformation.}
Our model is trained on a synthetically-generated dataset.
We validate that it generalizes to real user edits using a prototype interface illustrating our segment-based editing workflow.
The user can segment any part of the image and transform, duplicate, or delete it. 
We provide a video demonstrating this editing interface in the supplementary materials, and contrast the user's edit with the model's output.
To the best of our knowledge, no previous work focuses exactly on our use case (photorealistic spatial edits), so we adapt closely related techniques to our problem setting for comparison.
Specifically, we compare to the following baselines:
\begin{enumerate}
    \item SDEdit \cite{meng2022sdedit}: a general text-based editing method that trades off the adherence to the input image and the faithfulness to the text. This is the most general method we compare against, as we can directly provide it with the coarse user edit and a generated caption. Since SDEdit can take the coarse edit directly as an input, we emphasize it the most in our comparisons.
    % \item AnyDoor \cite{chen2023anydoor}: an image compositing model that harmonizes objects from a source frame to a target frame. We follow the author's method of using it for spatially compositing an image by inpainting the object using an off-the-shelf inpainting algorithm and re-inserting the object into the desired location.
    \item DragDiffusion \cite{shi2023dragdiffusion}: a drag-based editing model that takes source-target key-handles to move parts of the object for re-posing.
    \item InstructPix2Pix \cite{brooks2023instructpix2pix}: a text-based editing method that follows an instruction-like style captions.
    \item MasaCtrl \cite{cao_2023_masactrl}: a text-based editing method that achieves a consistent identity preservation through a self-attention mechanism. However, it relies on DDIM inversion as an essential step, which can compromise the method's robustness.
    %

    %
    % \item MotionGuidance \cite{geng2024motion}: flow-based TODO???
\end{enumerate}

\topic{Adapting the baselines.}
We convert our inputs to the interface expected by these baselines for comparison.
SDEdit requires choosing a strength parameter dictating the amount of noise added to the input and trades off between faithfulness and unconstrained synthesis.
We set the strength to 0.4 in all experiments, i.e. we start at 40\% of the way through the diffusion process, adding the corresponding level of noise to $I_\text{coarse}$.
Unlike ours, their model expects a text input, which we automatically compute using BLIP~\cite{li2022blip}. We use the same generated caption with the other text-based methods (MasaCtrl and InstructPix2Pix) for the quantitative evaluation on our large corpus. For the qualitative evaluation on user edits, we choose a caption that describes the user edit to the best of our abilities. However, we note that text description of spatial edits is inherently ambiguous (which is the motivation of our proposed method).
%
% To insert an object into a scene with AnyDoor, the user selects the object in a source image, and the destination region in a different target image. 
% %
% To adapt it to our use case, we follow the authors' suggestion of using the same image as source and target, using an off-the-shelf inpainting model to remove the selected object, then re-inserting it in a different image region. 
% %
% Their method offers limited control: the size of the insertion region is the only way to control the synthesized pose. 

To compare with DragDiffusion~\cite{shi2023dragdiffusion}, we record the segment motion in our user interface, 
compute the motion vectors for each pixel, and use this information to automatically create the keypoint-handles input needed by DragDiffusion.

\begin{figure*}[!bht]
\centering
\includegraphics[width=\linewidth]{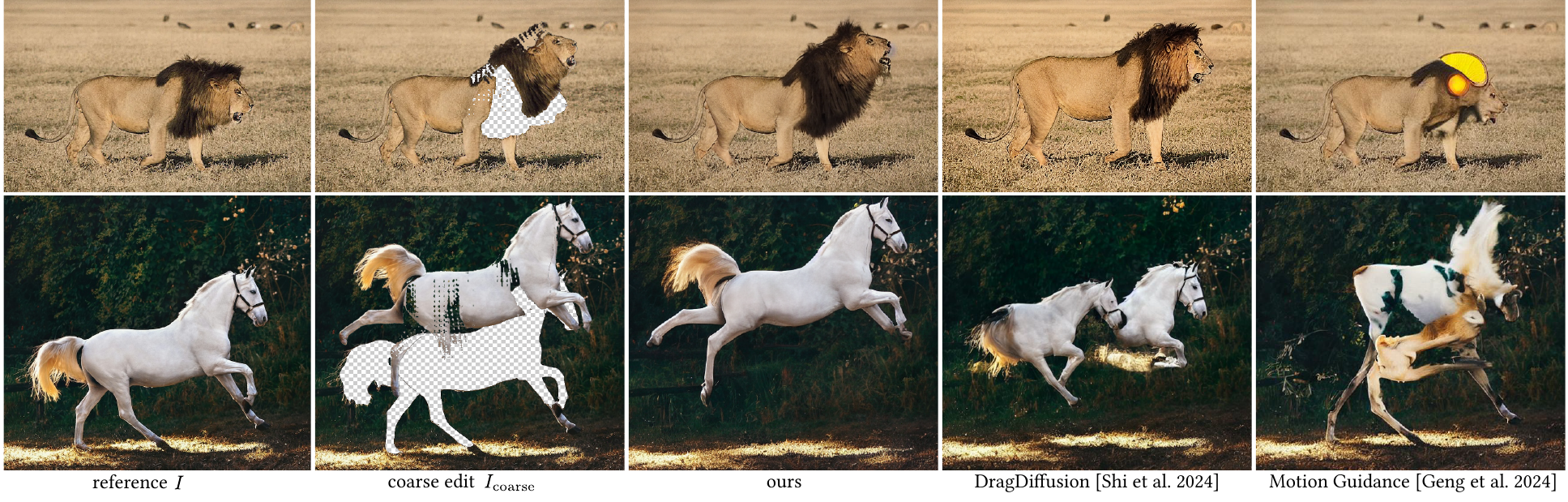}
\vspace{-5mm}
\caption{
\textbf{Comparison with DragDiffusion.} 
We use the Drag Diffusion ~\cite{shi2023dragdiffusion} to generate the results in the right column.
We seed dragging control points this method expects for each of the modified image segments, and displace them using the same affine transform used to produce our coarse edit (second column).
DragDiffusion generates fairly conservative image edits, and collapses with more drastic reposing edits.
However, our method successfully handles wide range of reposing levels.
Photos sourced from \copyright Unsplash.
}

% Two pipelines approach. Parallel pipeline to extract features from the original image, and injects attention using correspoendences and cross-reference attention. Using DINO for feature extractions. Finetune the entire model.

% \vspace{-10mm}

\label{fig:reposing_comparison}
\end{figure*}

\subsection{Quantitative evaluation}

While the task of image editing is inherently subjective, a natural task is to evaluate our method on edits generated from videos using the motion models we discuss earlier. We use a held-out split of our dataset, and evaluate our method against the baselines on the performance at reconstructing the target frame. In Table \ref{tab:quant_eval} we show that our method significantly outperforms the baselines on reconstruction metrics. We also computed the flow error, by comparing the RAFT optical flow between the reference and target, and the flow between the reference and the output. Intuitively, the smaller the flow error, the more that the method is faithful to the user edit. We show that our method is significantly outperforming all the baselines across all metrics. The second best method is DragDiffusion, which is likely due to the fact that it accepts a form of spatial edit as an input (through drag handles), and the LoRA optimization on each image to preserve identity. Across all text-based methods, SDEdit performs the best in general likely due to the fact that it can directly accept the user edit as an input rather than purely relying on the caption to perform the edit.

\begin{figure*}[!tbh]
% \begin{center}
\centering
\includegraphics[width=\linewidth]{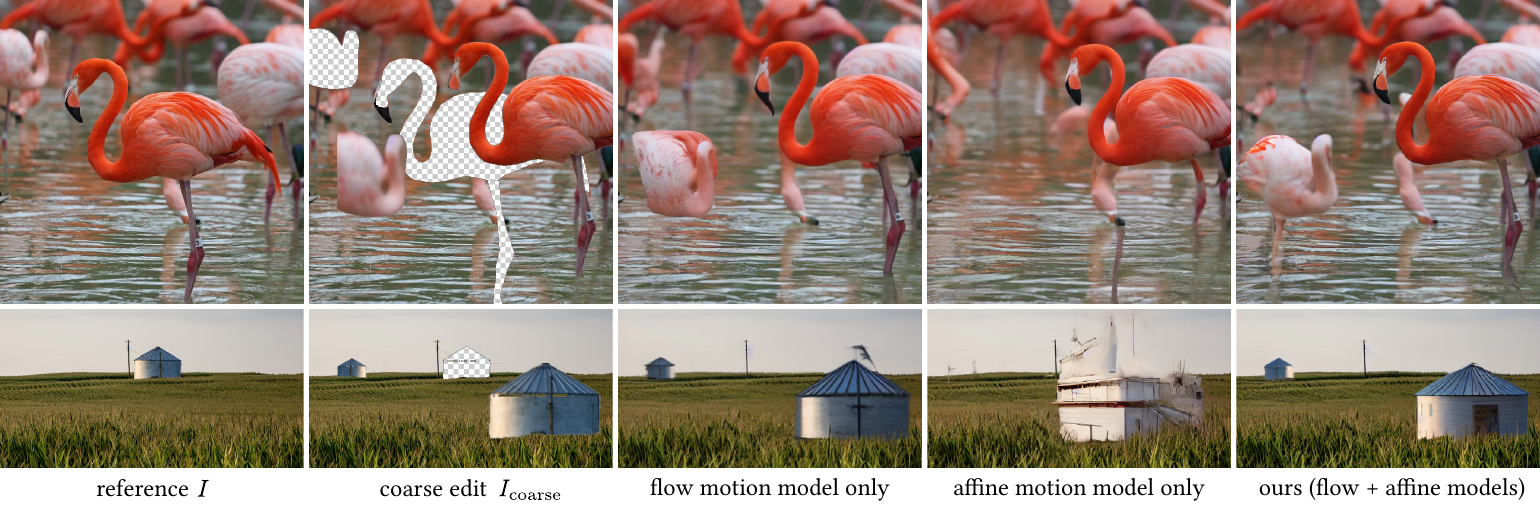}
\vspace{-5mm}
\caption{\label{fig:motion_model_ablation}
\textbf{Motion models ablation.} We compare how the 2 motion models we use to create our coarse edits (column 2) during training affect the model's behavior.
If we warp the reference frame (column 1) using the flow only (column 3), the model learns how to harmonize the edges of the edited regions, but remains very conservative and does not add much details to increase realism.
On the other extreme, if we only use the piecewise affine motion model (column 4), the model learns to hallucinate excessively, losing its ability to preserve object identity.
Our full solution trains with both motion models (column 5) to increase the model versatility, allowing the model to generate realistic details while still maintaining good adherence to the user input.
Photos sourced from \copyright Unsplash.
}
% \vspace{-10mm}
% \hadi{need to highlight that the three on the right are outputs, and what we describe is the motion model they are being trained on.}}
\end{figure*}

\begin{figure*}[ht]
\centering
\includegraphics[width=\linewidth]{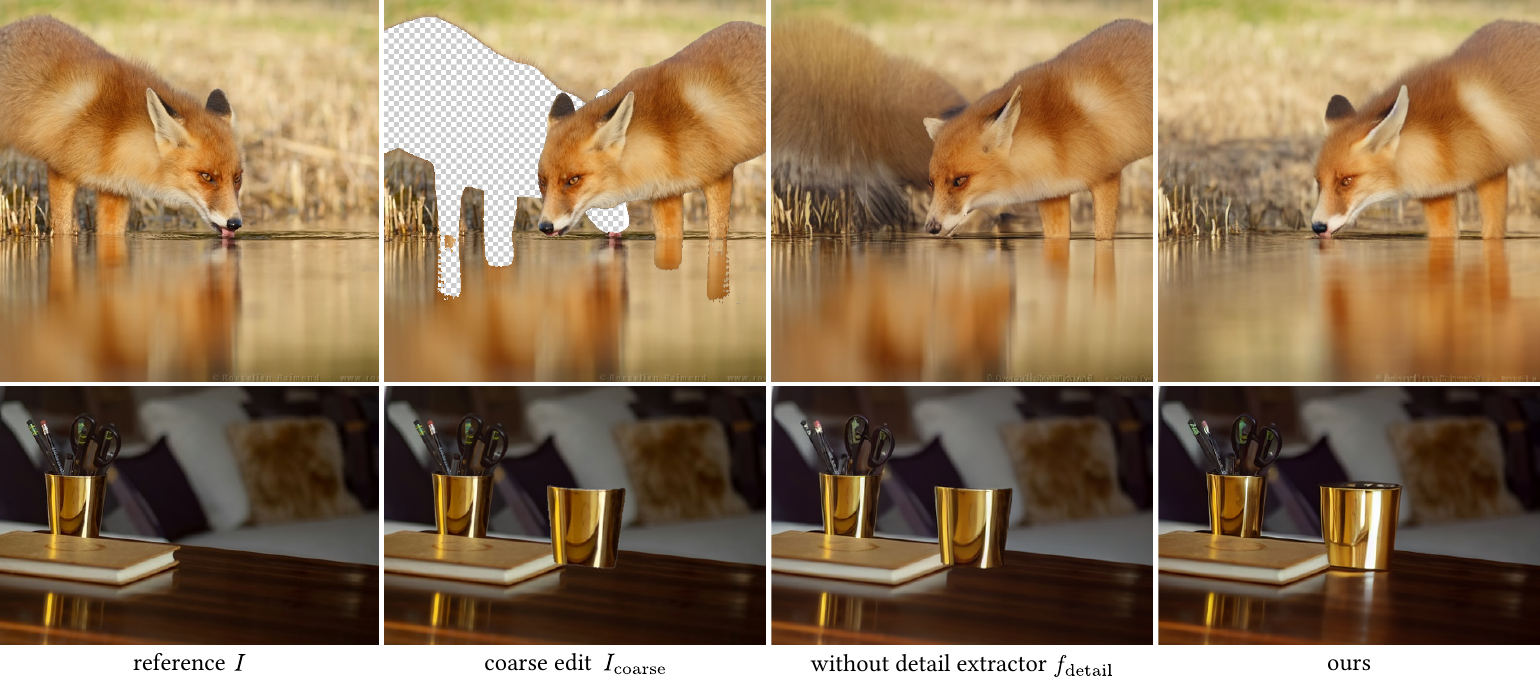}
% \mpage{0.23}{\frame{\includegraphics[width=1.0\linewidth, trim=0 0 0 0, clip]{figure_imgs/new_application_images/fox_drinking_og.png}}}\hfill
% \mpage{0.23}{\frame{\includegraphics[width=1.0\linewidth, trim=0 0 0 0, clip]{figure_imgs/new_application_images/fox_drinking__edit__018.png}}}\hfill
% \mpage{0.23}{\frame{\includegraphics[width=1.0\linewidth, trim=0 0 0 0, clip]{figure_imgs/crossattn_ablations/fox_drinking__edit__018_output.png}}}\hfill
% \mpage{0.23}{\frame{\includegraphics[width=1.0\linewidth, trim=0 0 0 0, clip]{figure_imgs/new_application_images/fox_drinking__edit__018_output.png}}}\hfill
% \mpage{0.23}{\frame{\includegraphics[width=1.0\linewidth, trim=0 0 0 0, clip]{figure_imgs/applications_images/kingfisher_og.png}}}\hfill
% \mpage{0.23}{\frame{\includegraphics[width=1.0\linewidth, trim=0 0 0 0, clip]{figure_imgs/applications_images/kingfisher__edit__003.png}}}\hfill
% \mpage{0.23}{\frame{\includegraphics[width=1.0\linewidth, trim=0 0 0 0, clip]{figure_imgs/crossattn_ablations/kingfisher__edit__003_output.png}}}\hfill
% \mpage{0.23}{\frame{\includegraphics[width=1.0\linewidth, trim=0 0 0 0, clip]{figure_imgs/applications_images/kingfisher__edit__003_output.png}}}\hfill
% \mpage{0.23}{reference} \hfill \mpage{0.23}{coarse edit $I_{coarse}$} \hfill
% \mpage{0.23}{without detail extractor $f_{detail}$} \hfill \mpage{0.23}{ours (full method)}
\vspace{-5mm}
\caption{\label{fig:architecture_ablation}
\textbf{Architecture ablation.} Without the detail extractor branch and using CLIP to extract the reference features (3rd column), the model struggles with spatial reasoning as it cannot access the grounding of the original reference image (1st column).
This ablation's outputs are overly conservative, not steering too far away from the coarse edit (2nd column).
Our full model produces much more realistic edits (4th column), with harmonious shadows and object-background contact.
It refines object boundaries and synthesizes plausible reflections. 
% does not ``hallucinate'' unrelated content when inpainting (bottom row).
Photos sourced from \copyright Roeselien Raymond and \copyright Unsplash.
}
% \vspace{-5mm}
\end{figure*}

\begin{table}[ht]
\caption{
\textbf{Quantitative evaluation.} Perceptual loss, SSIM, and flow error on a held-out validation subset of our video dataset. Our method significantly outperforms all the baselines across all metrics.
}
\centering
\footnotesize
\begin{tabular}{lcccc}
\toprule
\textbf{Method} & \textbf{Motion Model}  & \textbf{LPIPS $\downarrow$} & \textbf{SSIM $\uparrow$} & \textbf{Flow (px) $\downarrow$} \\
\midrule
MasaCtrl \cite{cao_2023_masactrl}  & - & 0.44  & 0.49 & 23.1 \\
 \midrule
IP2P \cite{brooks2023instructpix2pix}  & -  & 0.47 & 0.42 & 24.9 \\
\midrule
DragDiff$^*$ \cite{shi2023dragdiffusion}
\footnotemark
& Piecewise affine  & 0.26 & 0.58 & 16.9 \\
 & Flow-based  & 0.26 & 0.59 & 15.8 \\
\midrule
SDEdit \cite{meng2022sdedit}  & Piecewise affine  & $0.40$ & $0.57$ & $24.7$   \\
 & Flow-based  & $0.37$ & $0.61$ & $22.4$ \\
 \midrule
Ours & Piecewise affine   & $\mathbf{0.21 }$ & $\mathbf{0.70}$ & $\mathbf{5.33}$ \\
 & Flow-based    & $\mathbf{0.16 }$ & $\mathbf{0.78}$ & $\mathbf{3.06}$  \\
 % \midrule
 % Input(coarse edit) & Piecewise affine   & $0.25$ & $\mathbf{0.75}$ & $\mathbf{5.22}$ \\
 % & Flow-based    & $0.19$ & $\mathbf{0.83}$ & $\mathbf{2.9}$  \\
\bottomrule
\end{tabular}
\vspace{-1em}

% }
% \end{wraptable}
% \vspace{-5mm}
\label{tab:quant_eval}

\end{table}
\footnotetext{\new{We find that DragDiffusion fails to produce an output on edits with large motion, so we restricted this evaluation to the subset where it is able to produce an output. Note that the ordering of the methods remains the same if we restrict all methods to the same subset.}}

\subsection{Evaluation on user edits}
\new{
The training dataset we use for MagicFixup allows for a host of different applications. In Fig.~\ref{fig:diverse_applications} we highlight 12 different example applications we generated using MagicFixup, which includes edits outside of the training data, like perspective edits, 3D transformations, and even colorization. In this section, we discuss edits generated using our Collage Transform interface and compare against pose-editing baselines. We further highlight additional applications beyond image recomposition in the appendix.}

\topic{Collage transform editing.}
\new{Using our user interface, we created a collection of edits that spatially recompose photos. In Fig.~\ref{fig:applications} we show how our model adds realistic details to objects moved to a region of sharper focus, snaps disconnected objects together, and resynthesizes shadows and reflections as needed.}
Another natural baseline for spatial recomposition is inpainting an object and reinserting it in the image. We use AnyDoor~\cite{chen2023anydoor} as the insertion method, and compare the recomposition result.
In Fig.~\ref{fig:insertion_comparison}, we used our model to delete the dog (and automatically remove the shadow), and then re-inserted the dog using AnyDoor.
The dog's identity underwent significant changes, and AnyDoor does not harmonize the composite with the ground.
It also does not completely remove the halo caused by the inpainting mask in the destination region.
In contrast, our model synthesizes a coherent output without discontinuity artifacts.
We also used AnyDoor in duplicating the snake in Fig.~\ref{fig:teaser}, and we show that AnyDoor has a loss of identity on the snake, while our method correctly introduces some defocus blur to adjust for the reference's shallow depth of field.
We compare against text-only editing methods in Fig.~\ref{fig:text_comparison}, and show that InstructPix2Pix~\cite{brooks2023instructpix2pix} only alters the apperance without following the spatial edit instruction prompt, and MasaCtrl~\cite{cao_2023_masactrl} completely loses the input identity due to the failure of DDIM inversion. 
In the appendix, we also show additional comparisons against text-only methods.

\topic{Image reposing.}
Since we allow the user to edit the image by selecting segments of arbitrary size, the user can re-pose objects by selecting sub-parts and applying an affine transformation on them, effectively animating the object. 
 We compare our method to DragDiffusion \cite{shi2023dragdiffusion} that uses drag handles, and Motion Guidance \cite{geng2024motion} that uses flow to guide the diffusion sampling to follow the user edit. To ensure a fair comparison, we keep track of dense pixel correspondence in our Collage Transform user interface for the user edit. Using the dense correspondence maps, we can directly generate the drag handles and flow inputs these baselines require. 
%
% Note that in the lion example, 
In Fig.~\ref{fig:reposing_comparison}, DragDiffusion moves the lion's body higher up, which loosely aligns with the user edit, but is inconsistent with the user's intent of only moving the head.
This example highlights how a non-interactive point-dragging interface can be at odds with the user's desired output, because it does not provide a good preview of what the model would generate before running it.
Our Collage Transform interface is more immediate, and our coarse edit aligns with the final output. On the other hand, despite having dense flow, Motion Guidance completely fails to follow the user edit as the test time optimization process is unreliable.
In the second example, DragDiffusion collapses, likely because the user input is complex and goes beyond a minimal displacement of the subject that it can handle, and Motion Guidance lifts the horse up in the air but fails to keep it in one piece.

Note that both DragDiffusion and Motion Guidance require a costly test-time optimization for each input. On NVIDIA A100, DragDiffusion takes approximately 2 minutes, and Motion Guidance takes nearly an hour for a single input. In contrast, our method only requires the feed-forward sampling, taking approximately 5 seconds. 

\begin{figure}[pht]
\centering
\includegraphics[width=\linewidth]{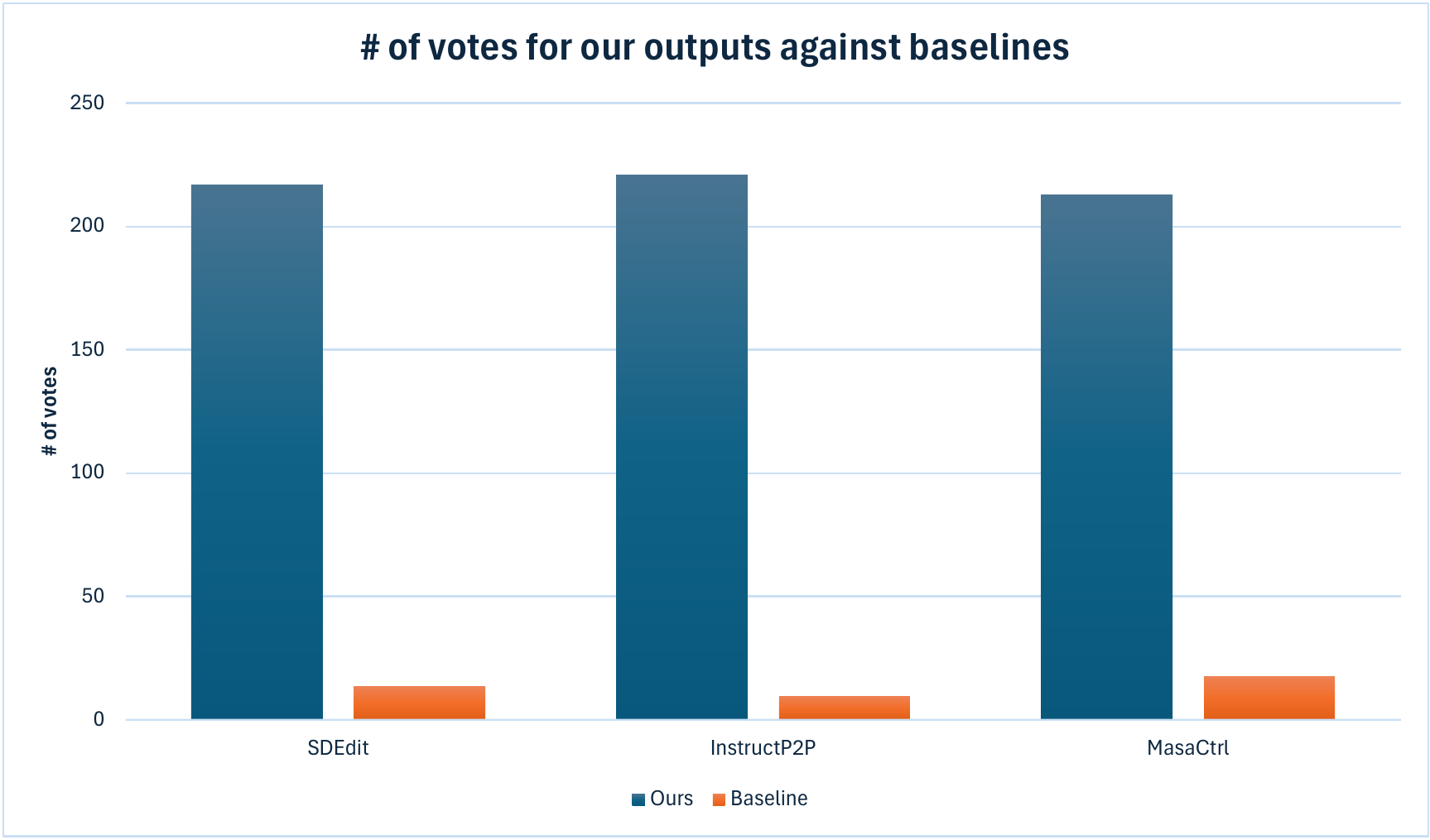}
\caption{\label{fig:user_study}
\textbf{User study results.}  We asked 21 users to compare the editing results from our method and the outputs using baselines. For each baseline, the user directly compares between our method's output and the baseline's. Overall, the users overwhelmingly preferred our method. Out of 231 votes, only 14 votes were for SDEdit \cite{meng2022sdedit}, 10 votes were for InstructPix2Pix \cite{brooks2023instructpix2pix}, and 19 votes for MasaCtrl \cite{cao_2023_masactrl}}
\vspace{-5mm}
\end{figure}
% \hadi{considering removing a row here so that it fits better with the figure above it.} \mg{I think it's fine. Maybe try to have this figure on the same page as the previous one "Ablating reference input", so avoid the odd break of references}

\begin{figure*}[!ht]
\centering
\includegraphics[width=\linewidth]{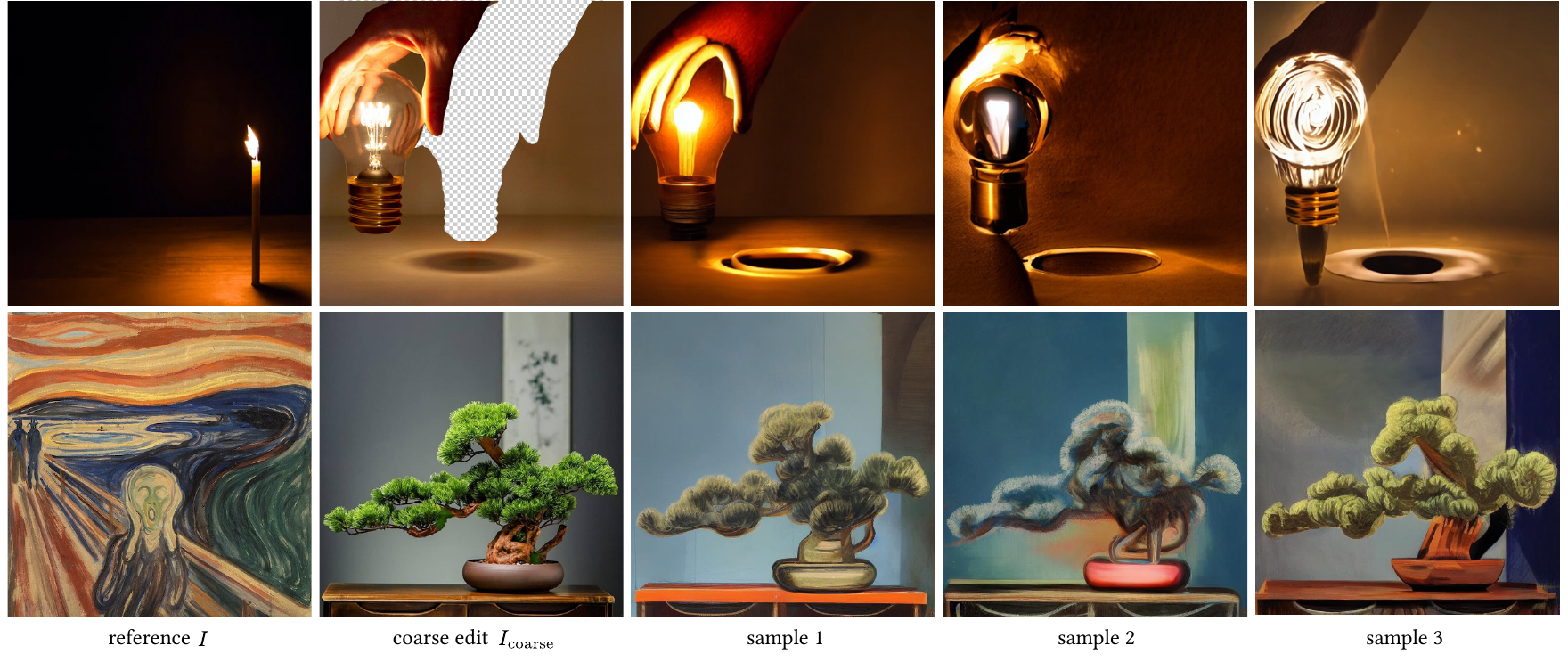}
\vspace{-5mm}
\caption{\label{fig:style_transfer}
\textbf{Ablating reference input.} When passing an arbitrary image as a reference to the detail extractor network independent of the input of the synthesis network, we find an effect similar to style transfer. The model preserves the spatial structure of the "coarse edit" input while maintaining the style and global appearance of the "reference." This behavior likely contributes to the model robustness in generalization to new domains and processing new types of edits. The scream painting is now in public domain, and remaining photos were sourced from \copyright Unsplash.
}
\end{figure*}

\topic{Perceptual user study.}
% \hadi{the rebuttal user study is thin on participants (14 users), so getting more users in the next few days and will update}
To evaluate the realism of our editing, we conducted a user study comparing the quality of our edits against three baselines: SDEdit \cite{meng2022sdedit}, InstrictPix2Pix \cite{brooks2023instructpix2pix}, and Masa-Ctrl \cite{cao_2023_masactrl}. We used 11 diverse photo edits, with 21 students participating and voting for all pairs of images. For each pair, we provided the users with the reference image as well as the \textit{intended} user edit, and asked for each sample the following ``For the following edit, which of those images do you find a more realistic result?'' in a 2-alternative forced-choice (2AFC) format. In Fig.~\ref{fig:user_study} we show the aggregated votes for our method against each baseline. We see that our method is overwhelmingly preferred. For each baseline comparison we have 231 total votes, and only 14 votes were for SDEdit \cite{meng2022sdedit}, 10 votes were for InstructPix2Pix \cite{brooks2023instructpix2pix}, and 19 votes for MasaCtrl \cite{cao_2023_masactrl}. We show comparison results against those baselines in Appendix~\ref{appendix:text_comparisons}, and we encourage the reader to compare the results directly.

%We include a more detailed analysis as well as the visuals used in the supplementary material.
% \mg{R3: show examples where our method is {\bf not} preferred.}
% \mg{R3: explain how the users are selected and whether they are related to authors}

\subsection{Ablation studies}
In this section, we evaluate the role that different motion models play, as well as the importance of cross-reference attention.
% we use as opposed to a simple img-CLIP based cross-attention.

\topic{Motion models ablation.}
Intuitively, training the model only on flow-warped images would prevent the model from learning to synthesize drastic changes, since flow-warping tends to be well-aligned around the edges. On the other hand, using the piecewise-affine motion model requires the model to adjust the pose of each segment (and learn to connect them together nicely), which forces the model to only use the input as a coarse conditioning. In Fig. \ref{fig:motion_model_ablation}, we show that the behavior of the model trained on different motion models is consistent with our intuition, where the model trained on flow-only preserves the content and refines the edges, while the model trained only on the piecewise-affine model struggles with preserving identity. On the other hand, the model trained on different motion models falls in the sweet-spot where it addresses user edits faithfully while adding content as needed. 

\topic{Architecture ablation.}
Prior work relies on using Image-CLIP embeddings or DINO features to encode the information of the content being inpainted or inserted into the image \cite{yang2023paint, chen2023anydoor}. The CLIP features are a reminiscent of the way Stable-Diffusion is trained with cross-attention with text CLIP embeddings. However, as we believe that CLIP features only carry semantic features that are too weak to pass useful information about the reference structure. We use a cross-reference mechanism, similar to Masa-Ctrl \cite{cao_2023_masactrl}, and unlike prior work, we completely remove the CLIP cross-attention layer. To validate our design decision, we compare using CLIP image embedding of the reference for cross-attention as opposed to the cross-reference-attention we propose. We observe that when relying only on CLIP embeddings, the model struggles in harmonizing the edited regions as shown in Fig.~\ref{fig:architecture_ablation}. We find that the ablated model is conservative, and cannot address secondary effects like reflections. 

\begin{table}[t]
% \renewcommand{\arraystretch}{0.7}
% \begin{wraptable}{r}{0.5\textwidth}
\centering
% \tiny
% \fontsize{5pt}{5pt}
%
\caption{
\textbf{Quantitative ablations.} Perceptual loss on a held-out validation set from our video dataset.
}
\footnotesize
% \resizebox{\linewidth}{!}{
\begin{tabular}{lcc}
\toprule
\textbf{Model \& Training Data} & \textbf{Test Data}  & \textbf{LPIPS $\downarrow$} \\
\midrule
Piecewise affine & Piecewise affine   & $\mathbf{0.231 \pm 0.007}$ \\
 & Flow-based   & $0.220 \pm 0.007$  \\
\midrule
Flow-based  & Piecewise affine  & $\mathbf{0.229 \pm 0.007}$  \\
 & Flow-based  & $\mathbf{0.190 \pm 0.007}$ \\
\midrule
Both motion models   & Piecewise affine & $0.327 \pm 0.007$  \\
 (no cross-ref attn) & Flow-based   & $0.269 \pm 0.008$   \\
\midrule
Both motion models  & Piecewise affine & $\mathbf{0.231 \pm 0.007}$  \\
 (Full method) & Flow-based & $\mathbf{0.196 \pm 0.007 }$  \\
\bottomrule
\end{tabular}
% }
% \end{wraptable}
% \vspace{-5mm}
\label{tab:motion_model}
\end{table}

\begin{figure*}[!htb]
\centering
\includegraphics[width=\linewidth]{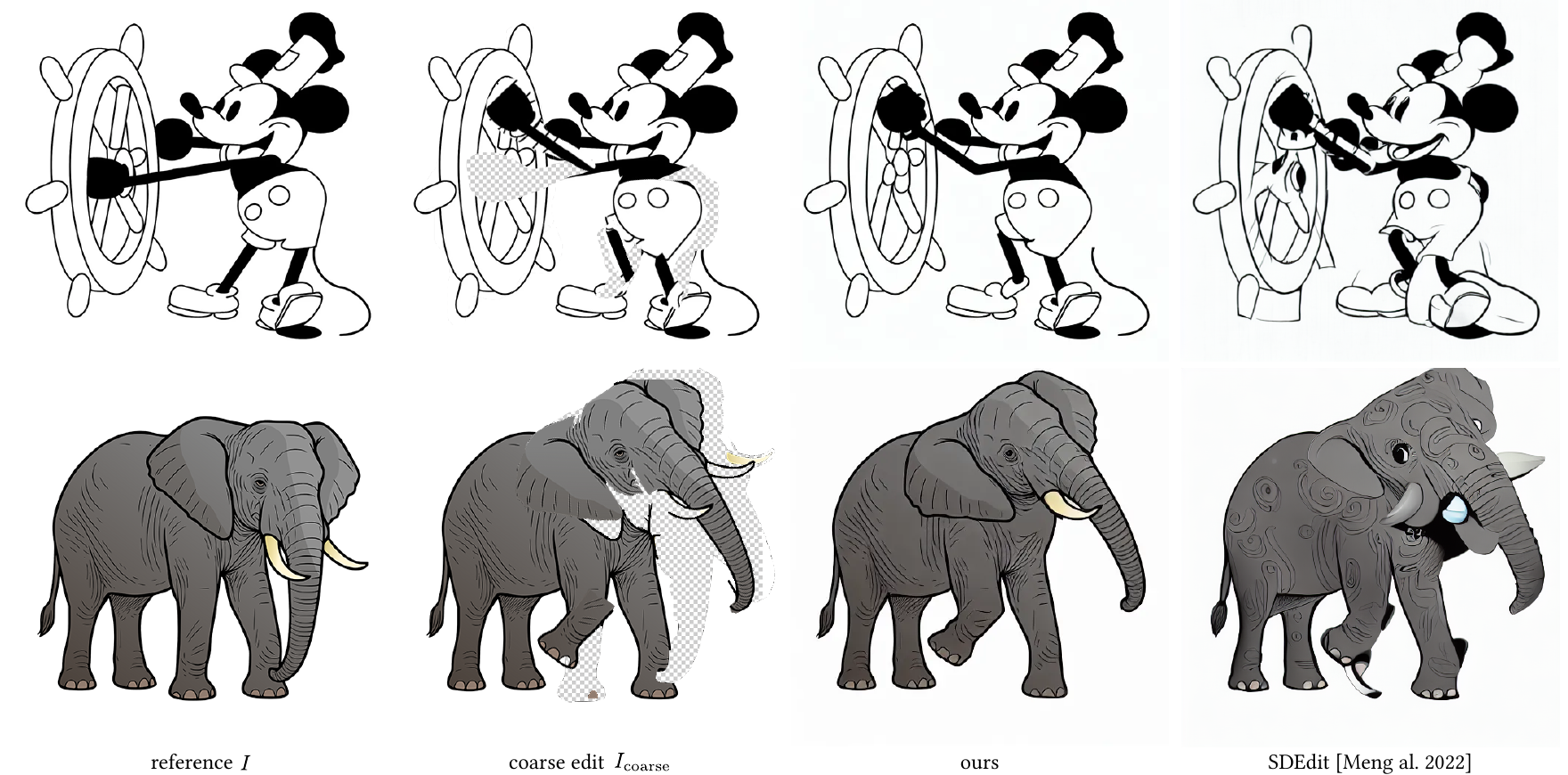}
\vspace{-5mm}
\caption{\label{fig:generalization}
\textbf{Beyond real photos.} 
Despite training the model on exclusively real videos, we find that the model can generalize to new domains beyond the training data, like cartoons and vector art. Photos are using public domain materials and CC licensed images.
% \hadi{considering removing a row here so that it fits better with the figure above it.} \mg{I think it's fine. Maybe try to have this figure on the same page as the previous one "Ablating reference input", so avoid the odd break of references}
}

\end{figure*}

\begin{figure*}[!hbt]
\centering
\includegraphics[width=1.0\linewidth]{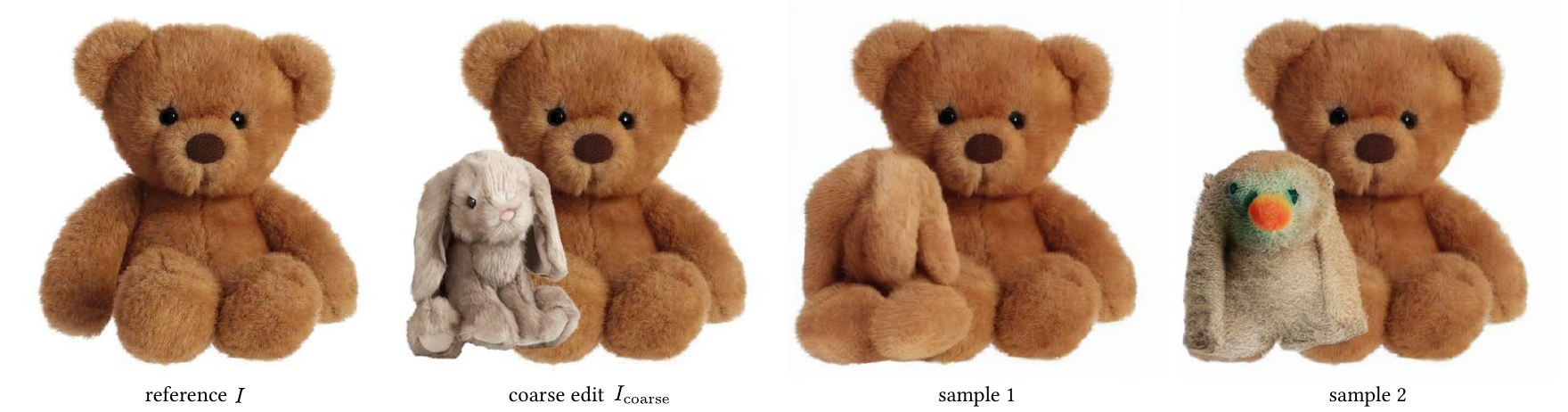}

\caption{\label{fig:limitations}
\textbf{Limitations}. Since the model can was only trained on spatial edits by rearranging the parts of a single image, the model struggles in inserting objects from outside the original image. Here we see that the model attempt to stylize the bunny to have an appearance similar to the teddy bear, but struggles to preserve the bunny's identity. Photos sourced from \copyright Unsplash.} 

% \vspace{-5mm}
\end{figure*}

\topic{Quantitative comparison.}
% \hadi{thinking of dropping this part. The table is not very informative when looking at the numbers}
% \mg{I am fine with leaving it in. The text is fairly short, and we got a lot of complaints for `not enough tables'. Are you worried that the numbers in the table are not very striking? Also ok to remove, if that's the reason.}
We evaluate our ablations on a held-out validation dataset from our video dataset. In the table on the right,
Table \ref{tab:motion_model}, 
we show that the model trained with flow-data and affine-motion are the top performers on perceptual loss on both types of test and that dropping the cross-reference attention and replacing it with CLIP embedding causes a severe drop in performance.

\topic{Detail extractor inputs}
To better understand how the model is utilizing the reference image in preserving the image details, we ask, what would happen if the provided "reference image" is completely independent from the provided "coarse edit?" Note that this case never occurs in training, but by modifying the inputs, we can gain insights on the inner workings of the model. In Fig.~\ref{fig:style_transfer} we show how the model behaves when provided an unrelated reference and edited images, and highlight diverse samples from the model. We notice that the model preserves the "style" and global appearance of the reference image, while preserving the spatial structure of the coarse edit. Intuitively, the reference image provides a sample of a clean image is supposed to look like, while the coarse edit provides the guidance of the spatial structure that the user intends to keep. This also touches on recent style transfer work \cite{Hertz2024stylealigned} that achieves style transfer through shared attention layers, which is similar to the effect we see here where the model transfers the style of the reference to the coarse edit. \new{In Appendix~\ref{sec:model_input_ablation} we provide additional quantitative analysis ablating the inputs of the detail extractor and synthesis UNets, like the disocclusion masks and the denoising timestep embedding.}

\subsection{Generalization beyond real photos}
While we only use video data of real videos, and filter out the majority of non-photorealistic videos in our training data, we explore the model's ability to generalize to completely new domains. In Fig.~\ref{fig:generalization} we show the model's ability to generalize to new domains, like cartoons and vector art.
We find that the model smoothly re-connects any disconnected parts of the image, and correctly re-synthesizes the art's outline that was lost in the editing process.
%We find that in the first example, the model properly connects the arms to the body, introduces clothe creases as needed, and also fix the depth ordering between the hand on the left and the leg. And in the last example, we find that the model correctly removes the shadow of the apple after removing it, but slightly falls short of adding a new shadow.
%
On the other hand, we find that SDEdit fails on all of those domains.
%We believe that the detail extractor network that takes the reference as an input plays a major role in the generalization, as we showed in the ablation study that the model uses the reference image to guide the global appearance of the image, boosting the versatility and generalization of the model.

\section{Limitations and conclusions}
\label{sec:limitations}
\label{sec:conclusions}
We present a method of assisting artists in photo editing through generative models while retaining a large level of control that traditional editing pipelines provide. 
We observe that with the appropriate motion model, we can use videos to train a model that can serve as a direct plugin in the editing process. 
We hope that our work inspires future editing research that can simply remove the cumbersome last-mile work by the press of a button.

Our generative model is trained for spatial compositions using video data.
It can spatially re-compose parts of the image but would struggle to insert objects from a completely different image as shown in Figure.~\ref{fig:limitations}.
Furthermore, we inherit the limitations of Latent Diffusion Models, which we use as our base models, especially for generating hands, faces, and small objects. 

\paragraph{Acknowledgement.} We would like to thank Sachin Shah, Nick Kolkin, Paul Guerro, Bryan Russell, and Tianwei Yin for fruitful discussions and feedback in revising the paper.

% \mg{R1 show visual examples of limitations (in supp)}
% TODO UNCOMMENT

% \hadi{citations seem to not be in ToG format even after changing it to ACM-reference-format?}
% \mg{I think that's fine for the review.}
\bibliographystyle{ACM-Reference-Format}
\bibliography{main}

\newpage
\
 \setcounter{section}{0}
\renewcommand{\thesection}{\Alph{section}}
% TODO UNCOMMENT
\section{Appendix: Additional applications}
We trained MagicFixup on inputs edited using affine transforms and flow warping. 
In this section, we explore how the model works under different types of edits, to better understand how the model function and cleans up different types of coarse edits. 
We find that the model is surprisingly robust and consistently produces photorealistic outputs, even in extreme edits like colorization that were completely different from our training data.

\begin{figure*}[hb]
\centering
\includegraphics[width=1.0\linewidth]{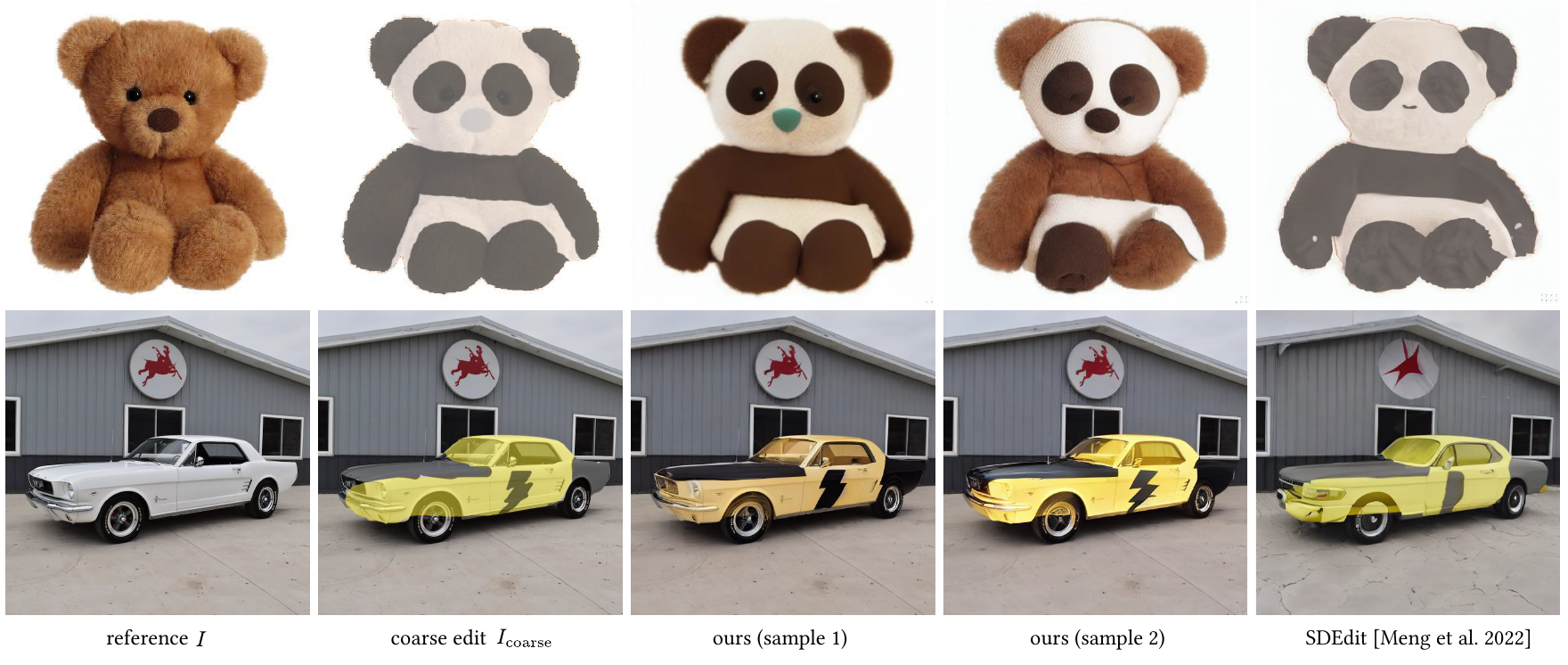}
\caption{\label{fig:colorization}
\textbf{Colorization.} Even though our motion models only included spatial transformations, we experiment with running the model on out of domain coarse edits. Surprisingly, we find the model to synthesize realistic colorized outputs. The model also cleans up uneven coarse edges. For example, the lightning drawn on the Mustang contains uneven curvy edges, and the model cleans it up nicely. We also show multiple samples to highlight the diversity of the outputs the model can generate to address these edits. Photos sourced from \copyright Unsplash.
}
% \vspace{-5mm}
\end{figure*}

\topic{Colorization}
Out model was only trained to enable spatial edits. So, we do not expect it to work well on significantly different image editing tasks like changing an object's color: our model's inputs only provide spatial transform information.
We tried to see what happens nonetheless.
In this task, we coarsely edited an object's color, before passing this coarse edit to our model. We show the results in Fig.~\ref{fig:colorization}
To our surprise, we found that our model can generate reasonable re-colorings!

\begin{figure*}[hb]
\centering
\includegraphics[width=1.0\linewidth]{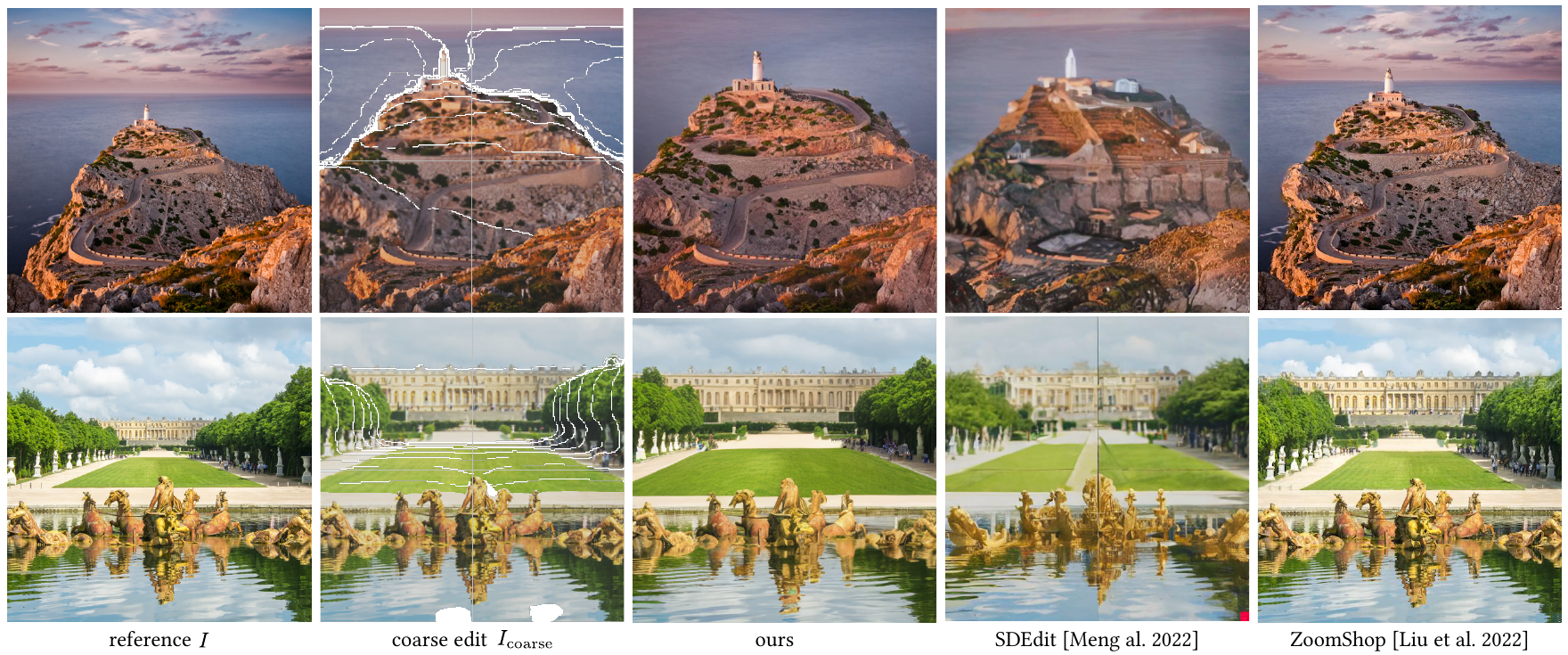}
\caption{\label{fig:zoomshop}
\textbf{Perspective editing.} By unprojecting the scene, then reprojecting different regions using variable camera parameters, we can manipulate perspective and make distant objects appear larger. While it is time consuming to create a high quality perspective edit (ZoomShop \cite{zoomshop} takes as long as 4 hours of manual labor), by using MagicFixup we can take a coarse reprojection and make it realistic. Here we attempt to reproduce the results from ZoomShop with our method, and include their results as a point of reference.
Photos sourced from ZoomShop \cite{zoomshop}.
}
% \vspace{-5mm}
\end{figure*}

\topic{Perspective editing}
The task of perspective editing involves warping the image to simulate capturing different parts of image with different focal lengths. 
Previously, Zoomshop \cite{zoomshop} achieves perspective editing by estimating the depth map of the image, unprojecting, then reprojecting different depth ranges. 
The warping operation creates holes, which are then inpainted using an off the shelf method.
However, a critical limitation of Zoomshop is that it can take as long as 4 hours of manual editing to achieve a clean edit, as the authors state.
This is because the method requires a perfect depth map that is carefully edited by the user, as well as manually cleaning up the inpainting.
However, MagicFixup excels at cleaning up coarse edits, so we implemented our own perspective editing pipeline to test our model.
This pipeline lets the user unproject the scene into a set of Multi-Plane-Images \cite{single_view_mpi}, and reprojects each plane using the user's desired field of view.
In Fig.~\ref{fig:zoomshop} we show the results of MagicFixup for perspective editing.
We attempted to visually match the edits shown in ZoomShop, and include their results as a reference only (we do not have access to the original intermediate edits from ZoomShop).
We find that MagicFixup introduces a super-resolution effect when enlarging distant background objects like the hill in the background, and also seamlessly cleans up the holes and seams created from the reprojection.

\begin{figure*}[ht]
\centering
\includegraphics[width=1.0\linewidth]{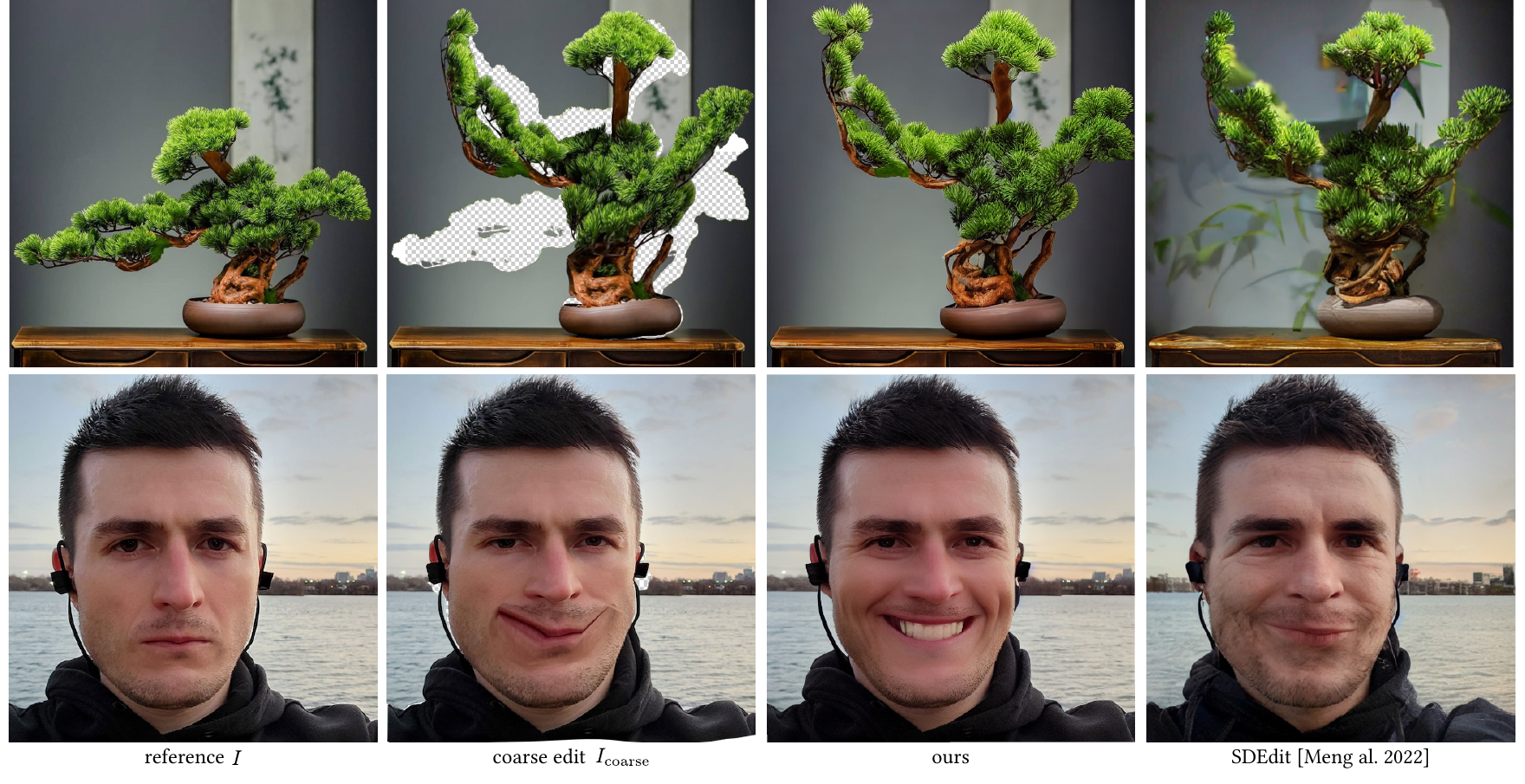}

\caption{\label{fig:photoshop}
\textbf{Photoshop's puppet warp.} We experiment with finer grain deformation using Photoshop's puppet warp feature. We deform the bonsai tree to resemble a dancing figure, and introduce a rough smile to a person's face. Our model then improves the realism of the edit. Note that in the second row, the model introduced a natural smile along with wrinkles around the mouth and eyes to display a more natural smile. Photos sourced from \copyright Unsplash and CC materials.
}
\end{figure*}

\topic{More complex deformations.} To allow for more complex spatial edits than simple affine transformations, we experimented with Adobe Photoshop's Puppetwarp tool to create more complex deformations.
In Fig.~\ref{fig:photoshop} we show results of a bonsai tree reshaped into a dancing-like figure.
We also applied Puppetwarp to edit a portrait to add a very coarse smile, and our model significantly improves the quality of the edit, even adding subtle face wrinkles associated with the smile, on the cheeks and around the eyes.

\begin{figure*}[hb]
\centering
\includegraphics[width=\linewidth]{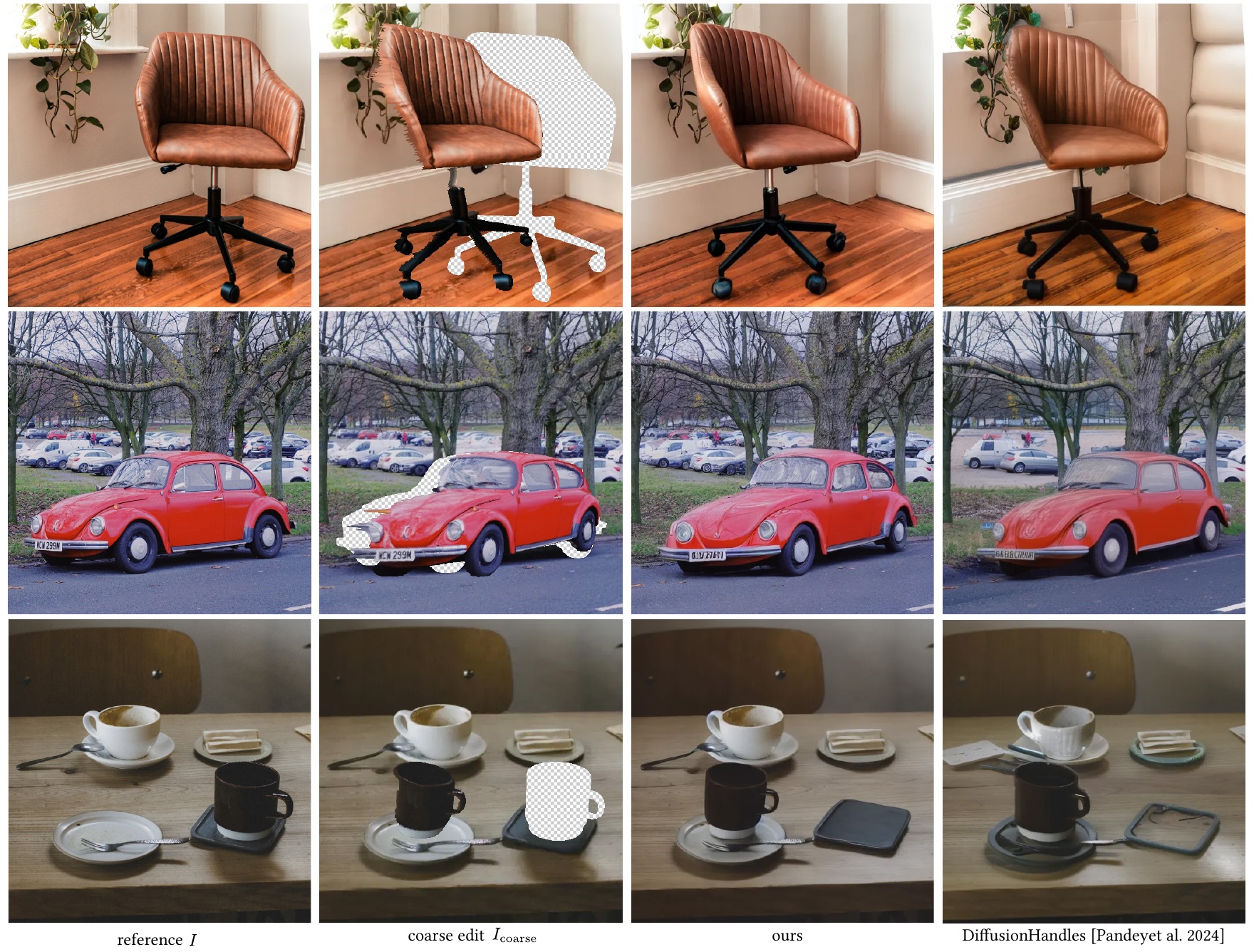}
\caption{\label{fig:transformation}
\textbf{3D transformations.} By unprojecting the image, applying 3D transformations on the unprojected point cloud, and reprojecting, we can achieve coarse 3D edits. We show that MagicFixup can addresses the artifacts generated from the reprojection, while preserving the image identity. On the other hand, we find DiffusionHandles \cite{pandey2024diffusionhandles} to alter the background identity on the right wall of the first example, the number of cars in the second example, and the shading of the plate and altering the identity of the spoon next to the white mug.
Photos sourced from \copyright Unsplash.
}
% \vspace{-5mm}
\end{figure*}

\topic{3D transformations.}
To apply 3D transformations, DiffusionHandles \cite{pandey2024diffusionhandles} proposes to use depth estimation to unproject the image, and applying 3D transformation on the object of interest followed by reprojection.
In DiffusionHandles, the authors use the transformed depth map as the input to a depth conditioned ControlNet \cite{zhang2023controlnet}.
We can directly use the (coarse) reprojected RGB as input to our model to enable similar 3D reprojections.
In Fig.~\ref{fig:transformation} we show 3D editing results using MagicFixup.
We show we can handle 3D transformations similarly to DiffusionHandles, and we outperform DiffusionHandles in preserving the identity of the image content.
In the first example, we see that DiffusionHandles alters the wall on the right, and in the second example it changes the number of cars parked in the background.
In the last row, the reflections on the mug in the background are altered, and the shading of the plate the mug was placed on became unnatural.
On the other hand, MagicFixup completely preserves object identity.
%
% We created 3D transformation edits
% %
% uses monocular depth to construct to create an unprojection then apply the 3D transformation to get the depth map of the edited image, and then  use a depth conditioned ControlNet \cite{zhang2023controlnet} to achieve the edit. To do 3D editing with MagicFixup, we also utilize monocular depth to unproject, then apply the 3D transformation, then reproject. This will give us a coarse 3D edited image, as shown in Figure \ref{fig:transformation}. We pass the reference and coarse edit to MagicFixup, and note that the model can help us achieve a 3D spatial edit in the image and do out of plane rotation. Notice how the shadow and reflections are addressed.

\section{Appendix: Reproducibility}

\begin{figure*}[ht]
\centering
\includegraphics[width=1.0\linewidth]{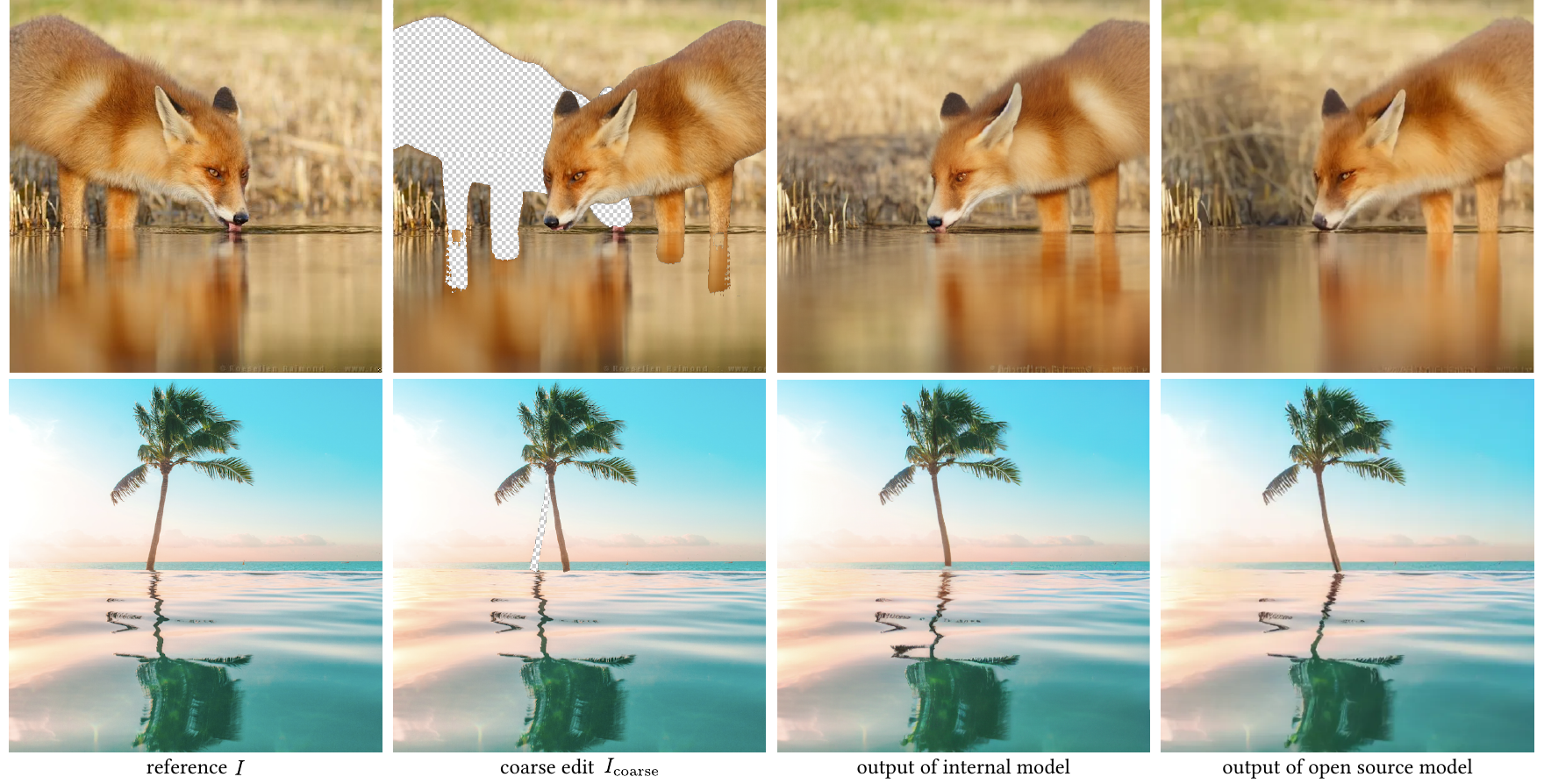}
\caption{\label{fig:open_source}
\textbf{Comparison with model trained on public data.} To maximize reproducibility, we train a version of the model on public video datasets that we plan to publicly release and open source. Here we show that the model trained on public data can similarly address secondary artifacts like reflections, and clean up artifacts due to coarse selection and editing as shown in the first row example with a coarse segmentation of the fox. Photos sourced from \copyright Roeselien Raymond \copyright Unsplash.
}
% \vspace{-5mm}
\end{figure*}

To ensure the reproducibility of our results, we plan to release our code along with a version of the model trained on a public video dataset. We use the public Moments in Time dataset (MiT) \cite{monfortmoments} for our open-source model, due to the similarity of the types of videos in our dataset, as our dataset contains stock-like clips similar to the ones in Moments in Time \cite{monfortmoments} and WebVid10M \cite{Bain21webvid}. We use 700k pairs of frames from MiT in contrast to 2.4M in the main model to train the model. We avoid using the larger WebVid10M dataset as it was recently taken down and the legality of using it is unclear. In Figure \ref{fig:open_source} we show that the open source model can achieve similar effects of addressing secondary artifacts like shadows and reflections.

\section{Appendix: Samples of our training data}
\begin{figure*}[!tbh]
% \vspace{-3mm}
\centering
\includegraphics[width=1.0\linewidth]{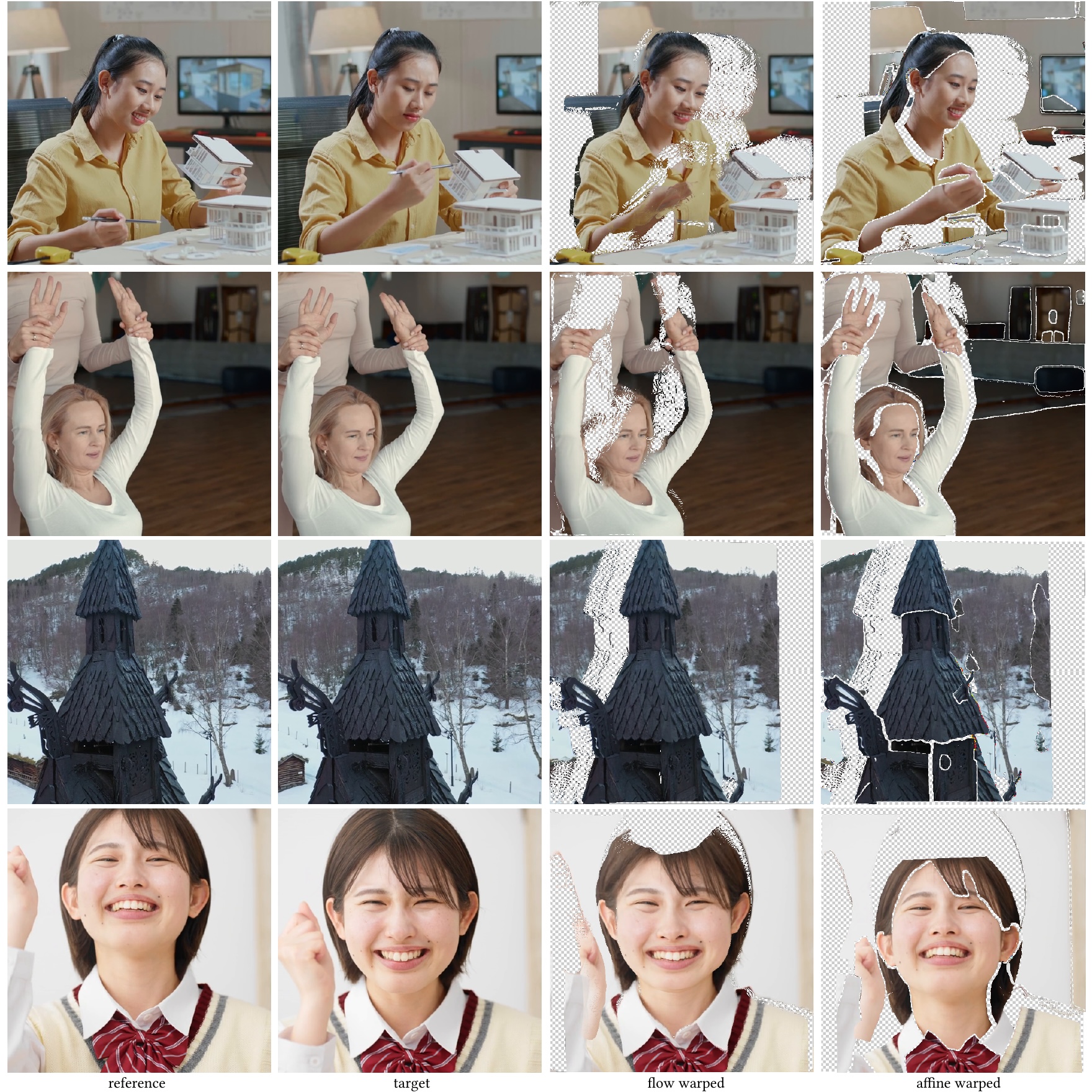}

\caption{
\textbf{Dataset samples.} We highlight some examples from our dataset along with the outputs of the flow and affine motion models. Note that the flow model densely aligns the reference image to the target, while the affine transformations provide much coarser alignment. For example, notice in the last row that in the flow warped image, the woman's smile and facial expression is aligned with the target, while in the affine warped we only see an alignment through scaling and shifting the person's segmentation mask. Media sourced from \copyright Adobe Stock.
}
% \vspace{-5mm}
% Two pipelines approach. Parallel pipeline to extract features from the original image, and injects attention using correspoendences and cross-reference attention. Using DINO for feature extractions. Finetune the entire model.
\label{fig:data_samples}
\end{figure*}

In Fig.~\ref{fig:data_samples} we show samples processed from our internal dataset that we used to train the model (show ref frame, target frame, flow warped, affine warped frames). Our videos come from stock-like internal dataset that is free of watermarks, unlike the commonly used public video datasets like WebVid10M \cite{Bain21webvid}.

\section{Appendix: Additional comparisons with text based methods}
\label{appendix:text_comparisons}
\begin{figure*}[!tbh]
% \vspace{-3mm}
\centering
\includegraphics[width=1.0\linewidth]{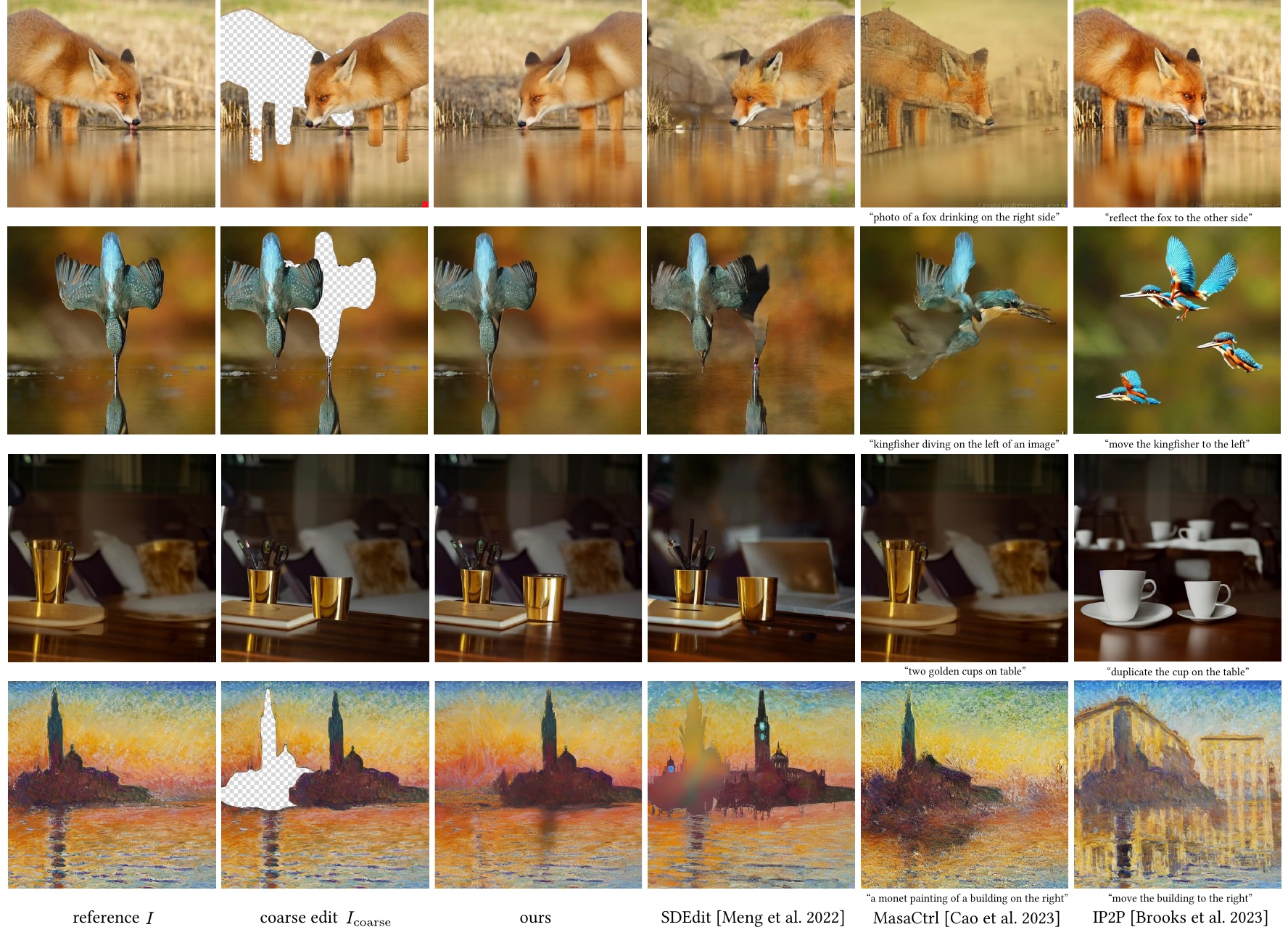}

\caption{
\textbf{Additional text comparisons.} We compare our method against text conditioned baselines MasaCtrl \cite{cao_2023_masactrl}
 and InstructPix2Pix \cite{brooks2023instructpix2pix}. To preserve the input identity, MasaCtrl requires a DDIM inversion step to the input image, which is prone to failing in reconstructing the input (as we show in the first two rows, the output identity is completely different from the input due to DDIM inversion failure), and in the cases where it succeeds in DDIM inversion, it is not possible to convey the user intent through a text prompt. Similarly, InstructPix2Pix either completely changes the identity of the image, or fails into editing the image to follow the text instruction. We show the text captions we used for both baselines under the image. We show the SDEdit \cite{meng2022sdedit} output as a reference, and we see that it is much more effective in following the user edit than the text baselines, which is why we rely on it as our main baseline.
 Photos sourced from \copyright Roeselien Raymond, \copyright Unsplash, and public domain data.
}
\label{fig:additional_text_comparison}
\end{figure*}

While text based editing methods lack the spatial control required to recompose photos precisely, we include additional qualitative comparisons for a comprehensive evaluation in Figure~\ref{fig:additional_text_comparison}. We compare against
InstructPix2Pix \cite{brooks2023instructpix2pix} and MasaCtrl \cite{cao_2023_masactrl}, and include our main baseline, SDEdit \cite{meng2022sdedit} for reference. To preserve input identity, MasaCtrl relies on a DDIM inversion step to reconstruct the input. However, inversion is not always reliable and can result in images that are similar in the high level appearance but with a different content from the input, as we show in the fox example. In contrast, we pass the input in a feed forward manner that allows the network to reliably preserve the input identity. On the other hand, InstructPix2Pix either leaves the input image intact with minimal changes, or severely alter the image identity, making it unreliable for spatial editing. We find that SDEdit is the most reliable baseline as it can take the user edit directly as an input, improving the spatial controllability. As a result, we use SDEdit as our primary baseline in the main paper.  

\section{Appendix: Ablation on models inputs}
\label{sec:model_input_ablation}
\new{For a comprehensive ablation study of the inputs to the detail extractor and  synthesis UNets. We analyze the role of the mask to the different UNets, and we experiment with dropping the timestep embedding from the detail extractor UNet. Dropping the time embedding in the detail extractor is equivalent to doing a one-time feature extraction, which would make it similar in style to using CLIP or DINO features rather than doing a feature extraction that depends on the current denoising step.
We finetune our main model for each of these configurations on images generated using WebVid-10M as we no longer have access to the original internal data. For fairness, we also finetune the main model on the same dataset. We show the results in Tab.~\ref{tab:ref_input_ablation}. Overall, timestep embedding in the reference UNet is essential. Providing the mask to the reference UNet is not needed, but the performance difference is within the standard error. \\
We find that providing the timestep embedding is essential for performance, which indicates that the detail extractor network extracts different features from the reference throughout the denoising process. This supports the intuitive understanding that the image generation process requires different levels of details for each step, as diffusion model generally starts by generating the coarse structure of the output and then synthesizing the higher frequency details later on.
For the disocclusion mask, we find that the providing the mask is essential to the synthesis network, but provides no additional information to the detail extractor network. We also ablate including the noisy reference in the detail extractor UNet, and find that its effect on performance is negligible as expected.
%We keep the noisy reference and the mask as inputs to the detail extractor simply to maintain an elegant symmetric architecture.
}

\begin{table*}[ht]
\caption{
\new{\textbf{Ablation on models inputs}. Here we ablate the inputs to the detail-extractor and synthesis UNets. We find that the providing the mask is essential to the synthesis network, but provides no additional information to the detail extractor network. We also find that providing the timestep embedding is essential for performance, which indicates that the detail extractor network extracts different features from the reference throughout the denoising process.}}

\centering
\footnotesize
\begin{tabular}{lcccccc}
\toprule
 \multicolumn{1}{c}{\textbf{Motion model}} & \multicolumn{3}{c}{Flow motion model}  & \multicolumn{3}{c}{Piecewise affine motion model} \\
\midrule
\multicolumn{1}{l}{\textbf{Method}} &  \textbf{LPIPS $\downarrow$} & \textbf{SSIM $\uparrow$} & \textbf{Flow (px) $\downarrow$} & \textbf{LPIPS $\downarrow$} & \textbf{SSIM $\uparrow$} & \textbf{Flow (px) $\downarrow$} \\
\midrule
w/o mask in either UNet & $ 0.277 \pm  0.01$  & $ 0.615 \pm  0.01$ & $\mathbf{35.554 \pm  2.14}$ &  $ 0.294 \pm  0.01$  & $ 0.596 \pm  0.01$ & $\mathbf{34.066 \pm  2.19}$  \\
w/o mask in detail ext.  & $\mathbf{ 0.187 \pm  0.01}$  & $\mathbf{0.715 \pm  0.01}$ & $\mathbf{35.682 \pm  1.88}$ & $\mathbf{ 0.220 \pm  0.01}$  & $\mathbf{0.667 \pm  0.01}$ & $\mathbf{36.051 \pm  2.09}$ \\
w/o timestep in detail ext. & $ 0.504 \pm  0.01$  & $ 0.498 \pm  0.01$ & $ 71.769 \pm  3.58$  & $ 0.579 \pm  0.01$  & $ 0.457 \pm  0.01$ & $ 73.004 \pm  3.49$  \\
 w/o noisy input in detail ext. & $\mathbf{ 0.207 \pm  0.01}$  & $\mathbf{0.699 \pm  0.01}$ & $\mathbf{37.654 \pm  2.03}$  & $\mathbf{0.242 \pm  0.01}$  & $\mathbf{0.655 \pm  0.01}$ & $\mathbf{37.126 \pm  1.99}$ \\
 ours   & $\mathbf{ 0.194 \pm  0.01}$  & $\mathbf{0.708 \pm  0.01}$ & $\mathbf{35.716 \pm  1.93}$  & $\mathbf{0.232 \pm  0.01}$  & $\mathbf{0.657 \pm  0.01}$ & $\mathbf{35.785 \pm  1.93}$ \\
\bottomrule
\end{tabular}
% \vspace{-1em}
\label{tab:ref_input_ablation}
\end{table*}

\section{Appendix: Quantiative evaluation using CLEVR}
\begin{figure*}[ht]
\centering
\includegraphics[width=1.0\linewidth]{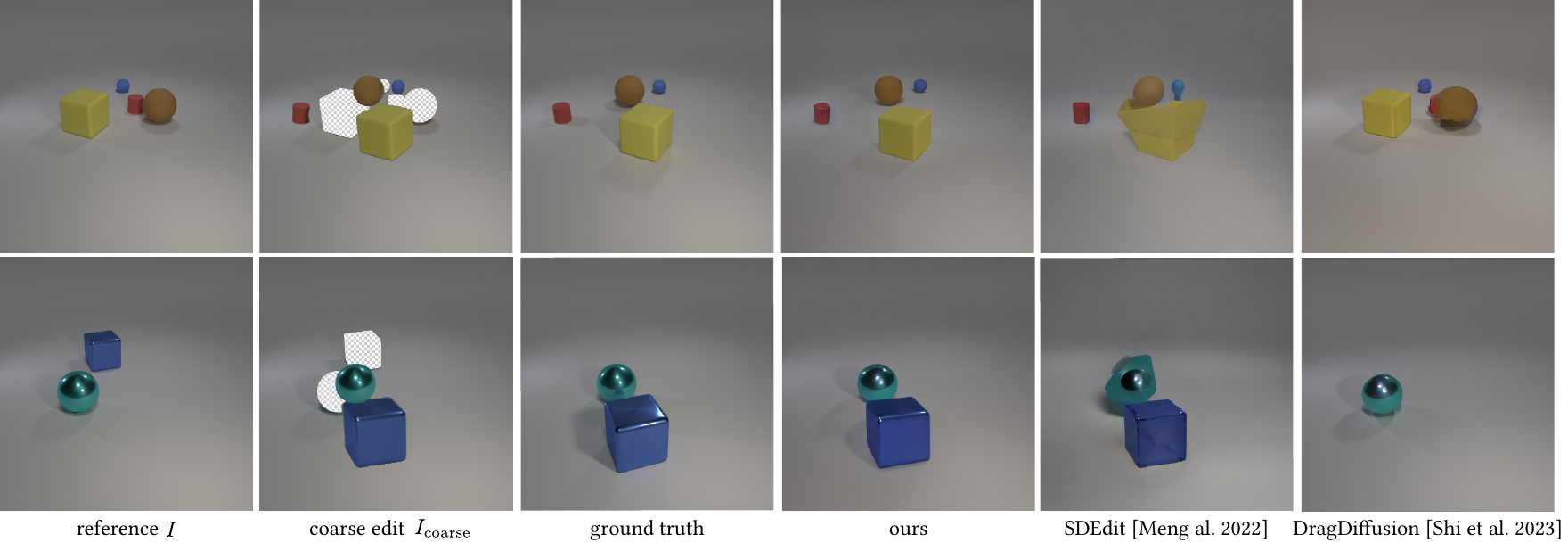}
\caption{\label{fig:clevr}
\textbf{CLEVR rearranging.} \new{We modify the CLEVR dataset generation pipeline to generate random sets of objects in different spatial arrangements, and synthesize a coarse edit that corresponds to re-aligning one arrangement to the other. Compared to the baselines, our method can better preserve the objects and harmonizes it with the environments. We find that Drag based methods like DragDiffusion struggle in generating motion beyond reposing, and SDEdit drastically morphs the objects.}
}
% \vspace{-5mm}
\end{figure*}

\new{
We used evaluation dataset generated through our dataset construction pipeline, as it provides a natural test set for spatial editing. It is challenging to construct large scale evaluation datasets without developing a novel motion model that we avoid training on. However, to substantiate our results further, we rely on the CLEVR dataset. CLEVR places objects with varying materials on a surface board, and includes multiple lights that showcase interesting shadows and shading. To levarage CLEVR to evaluate our model, we generate 50 random collection of objects, and rearrange each collection 3 times. Then, for an image in a given collection, we warp it to match the two other arrangements. This way we can automatically construct coarse edits, and have access to ground truth data for quantitative evaluation at the same time. In total, the test dataset consists of 300 edits. In Fig. ~\ref{fig:clevr}, we show two samples and the outputs of MagicFixup against SDEdit and DragDiffusion. We find that our method synthesizes new shadows and harmonizes the objects layering. On the other hand, while SDEdit can preserve the target arrangements, and DragDiffusion struggles to spatially relocate the objects. Since DragDiffusion cannot rearrange the objects, we restrict our quantiative evaluation in Table ~\ref{tab:clevr} to our method and SDEdit, and we find that our method outperforms the baseline in all metrics. While the CLEVR dataset is an imperfect test set, and out of distribution for our method, we find that the results further corroborate the robustness of our method and support our qualitative results.
}

\begin{table}[!htb]
\caption{
\textbf{CLEVR rearranging evaluation.} \new{We generate a version of the CLEVR dataset where synthesize 50 random collections of objects, and rearrange each collection in three different ways. We evaluate the method's performance in realistically rearranging the objects against the ground truth.}
}
% \centering
% \footnotesize
\begin{tabular}{lccc}
\toprule
\textbf{Method} & \textbf{LPIPS $\downarrow$} & \textbf{SSIM $\uparrow$} & \textbf{Flow (px) $\downarrow$} \\
\midrule
SDEdit  & $ 0.156 \pm  0.007$  & $ 0.886 \pm  0.002$ & $ 26.02 \pm  2.57$  \\
 % \midrule
Ours  & $\mathbf{0.078 \pm 0.007}$  &  $\mathbf{0.913 \pm 0.003}$ & $\mathbf{8.03 \pm  1.20}$ \\
\bottomrule
\end{tabular}
% \vspace{-1em}
\label{tab:clevr}
\end{table}

\section{Iterative editing}
\begin{figure*}[!htb]
\centering
\includegraphics[width=1.0\linewidth]{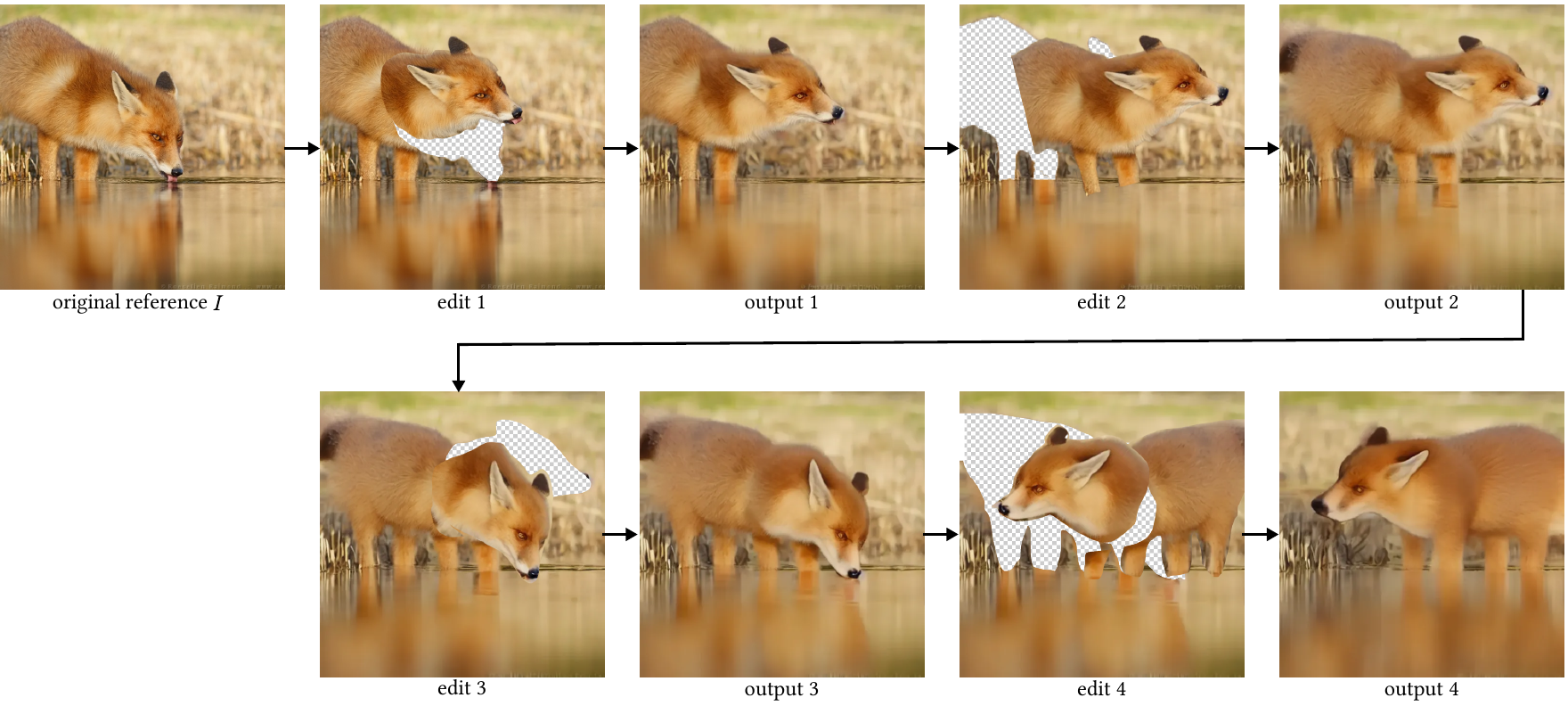}
\caption{\label{fig:iterative}
\textbf{Iterative editing.} \new{We iteratively edit the photo by setting the model's output as the new "reference" in each step, and spatially editing the model's output. Note that the model can coherently maintain the image's identity before it starts degrading with the fourth edit. We find that the iterative approach opens new possible edits. For example, in output 2 above, the model adds the rest of the fox's body, which provides context for additional edits that are not possible directly from the original image. Photo sourced from \copyright Roeselien Raymond.}
}
% \vspace{-5mm}
\end{figure*}

\new{One interesting question is how we can iteratively edit in image instead of making all the changes in one shot. In Figure  ~\ref{fig:iterative} we iteratively edited the photo of the fox next to the water in a manner that is almost similar to stop-motion animation. In each step, we apply the edit on the model's output from the previous step. So we set the model's output as the "reference" for the new edit.  We find that the model gracefully handles the first three iterative edits, and notice that the model's output. In the second edit, we see that the model auto-completes the body of the fox, and allows additional edits that are not possible with the original reference. We believe that our work paves a path for a future research direction that allows a human in the loop to interactively edit their photos alongside generative models.}

% \section{Coarse selection}

% \new{In our data generation pipeline, we used SegmentAnything to segment all objects in the image before applying affine transformations to the segment. As a result, we expect the model to generalize well to the selection level that matches SegmentAnything segments (which are not perfect). To investigate how the model degrades with coarse selections, we use Adobe Photoshop to compare accurate edge selection with rough lasso selection and coarse rectangle selection in Figure ~\ref{fig:coarse}. In the first row, with the table movment, we notice that the model gracefully handles lasso selection and nicely harmonizes the seams. With the rectangle selection, we find that the model also mostly harmonizes the output with a minor artifact on the left side that the model introduces. In the second example of the lioness in the savana, we find that the model blends the seams with all the selection mechanisms well. The results indicate that the model has a considerable tolerance to coarse selection.}
% \input{figures/revision_figures/coarse_edits}

\section{Appendix: Collage transform interface}
\begin{figure*}[ht]
\centering
\includegraphics[width=1.0\linewidth]{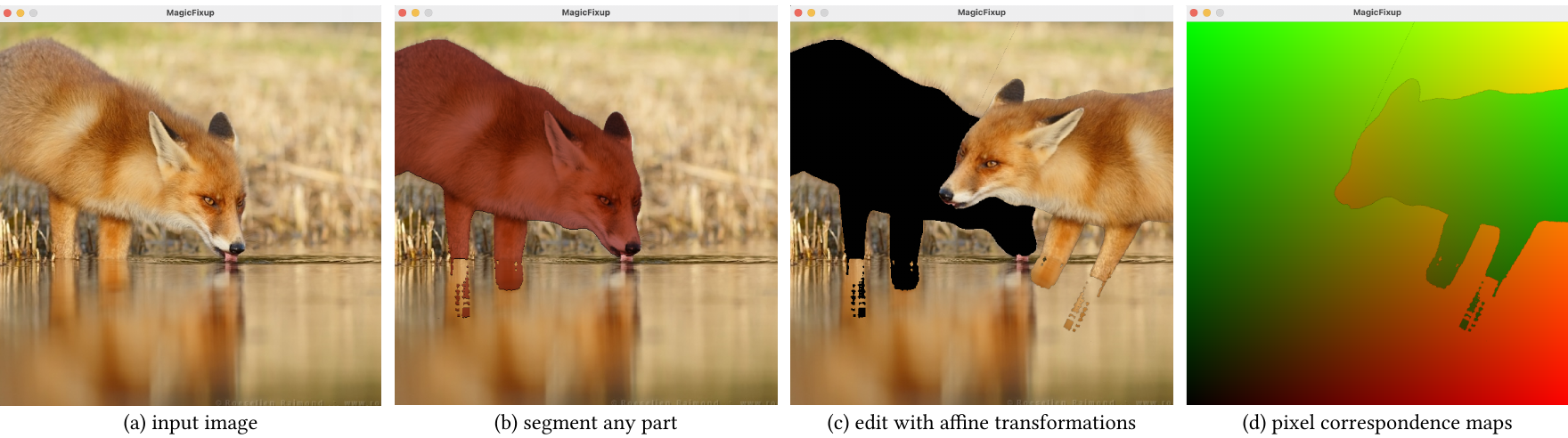}

\caption{\label{fig:collage_transform}
\textbf{Collage transform interface.} To create user edits while maintaining correspondences between the original image and the edit, we created the Collage transform interface. The user can select any object or part they would like to edit, apply the desired affine transformation, duplication, or deletion. The correspondence map that we maintain allow us to accurately and fairly compare against the baselines by computing flow or drag keyhandles using the correspondence maps. Photo sourced from \copyright Roeselien Raymond.
}
% \vspace{-5mm}
\end{figure*}

To facilitate creating user edits quickly, we created our own interface that supports the user selecting any object or parts they would like to edit, and make the edit by apply an affine transformation, duplication, or deletion. We show our user interface with an example demonstrating its usage in Figure.~\ref{fig:collage_transform} The interface allowed our users to create edits smoothly without any prior editing experience. Several of the edits used in this paper were created by novice users with no editing background.
Beyond the simplicity of the interface, we maintain a dense correspondence map between the pixels in the original image and the edit. The correspondence maps are critical to fairly compare against the baselines that take drag handles or dense flow as an input, as we can directly use the correspondence to compute the needed input.

\section{Appendix: Expanded user study with SDEdit}
% \hadi{the rebuttal user study is thin on participants (14 users), so getting more users in the next few days and will update}
Since SDEdit \cite{meng2022sdedit} is our primary baseline, we conduct an additional study only comparing our method with SDEDit, and significantly more user edits. We used 30 diverse photo edits, with 27 students participating and voting for all pairs of images. We conducted the study similar to the user study described in Section \ref{sec:result}, in a 2-alternative forced-choice (2AFC) format. For 80$\%$ of the edits, at least 75$\%$ of the users preferred our method. For the remaining images, except for one image, users preferred our method $65-80\%$ of the time. For one image in out of domain edit (editing a non-realistic artistic painting), users preferred both edits almost equally likely (52 $\%$ of users preferred SDEdit).

% TODO UNCOMMENT
% \input{sections/08_response}
\end{document}